\documentclass[journal]{IEEEtran}

\IEEEoverridecommandlockouts
\usepackage{cite}
\usepackage{float}
\usepackage{amsmath,amssymb,amsfonts,dsfont}
\usepackage{url} \usepackage{hyperref}

\usepackage{algorithmic}

\usepackage{graphicx}
\ifCLASSOPTIONcompsoc
\usepackage[caption=false, font=normalsize, labelfont=sf, textfont=sf]{subfig}
\else
\usepackage[caption=false, font=footnotesize]{subfig}
\fi
\usepackage{amsthm}
\theoremstyle{remark}

\newtheorem{example}{Example}

\usepackage{hyperref}
\hypersetup{
	colorlinks=true,
	linkcolor=blue,
	filecolor=magenta,      
	urlcolor=blue,
}
\usepackage{textcomp}

\usepackage{xcolor, soul}
\definecolor{lemonchiffon}{rgb}{1.0, 0.98, 0.8}
\sethlcolor{lemonchiffon}
\usepackage[inline]{enumitem}
\def\BibTeX{{\rm B\kern-.05em{\sc i\kern-.025em b}\kern-.08em
    T\kern-.1667em\lower.7ex\hbox{E}\kern-.125emX}}

\makeatletter
\newcommand{\linebreakand}{%
  \end{@IEEEauthorhalign}
  \hfill\mbox{}\par
  \mbox{}\hfill\begin{@IEEEauthorhalign}
}
\makeatother

\makeatletter %% <- make @ usable in command sequences
\newcount\SOUL@minus
\makeatother  %% <- revert @

\definecolor{comColor}{rgb}{0.08, 0.38, 0.74}

\begin{document}

\title{Reinforcement Learning-based Switching Controller for a Milliscale Robot in a Constrained Environment}

\author{\IEEEauthorblockN{Abbas Tariverdi\IEEEauthorrefmark{1},
Ulysse Côté-Allard\IEEEauthorrefmark{2}, Kim Mathiassen\IEEEauthorrefmark{4}, Ole J. Elle\IEEEauthorrefmark{2}\IEEEauthorrefmark{3}, \\ Håvard Kalvøy\IEEEauthorrefmark{5},  Ørjan G. Martinsen\IEEEauthorrefmark{1}\IEEEauthorrefmark{5}, Jim Tørresen\IEEEauthorrefmark{2}}\\
\IEEEauthorblockA{\IEEEauthorrefmark{1}Department of Physics,
University of Oslo, Oslo, Norway\\
\IEEEauthorrefmark{2}Department of Informatics,
University of Oslo, Oslo, Norway\\
\IEEEauthorrefmark{3}The Intervention Centre,
Oslo University Hospital, Oslo, Norway\\
\IEEEauthorrefmark{4}Department of Technology Systems,
University of Oslo, Oslo, Norway\\
\IEEEauthorrefmark{5}Department of Clinical and Biomedical Engineering,
Oslo University Hospital, Oslo, Norway
\thanks{Manuscript received November 7, 2021; revised December 24, 2022. 
Corresponding author: Abbas Tariverdi (email: abbast@uio.no).}}}

\maketitle

\begin{abstract} 

 This work presents a reinforcement learning-based switching control mechanism to autonomously move a ferromagnetic object (representing a milliscale robot) around obstacles within a constrained environment in the presence of disturbances. This mechanism can be used to navigate objects (e.g., capsule endoscopy, swarms of drug particles) through complex environments when active control is a necessity but where direct manipulation can be hazardous. The proposed control scheme consists of a switching control architecture implemented by two sub-controllers. The first sub-controller is designed to employ the robot's inverse kinematic solutions to do an environment search for the to-be-carried ferromagnetic particle while being robust to disturbances. The second sub-controller uses a customized rainbow algorithm to control a robotic arm, i.e., the UR5 robot, to carry a ferromagnetic particle to a desired position through a constrained environment. For the customized Rainbow algorithm, Quantile Huber loss from the Implicit Quantile Networks (IQN) algorithm and ResNet are employed. The proposed controller is first trained and tested in a real-time physics simulation engine (PyBullet). Afterward, the trained controller is transferred to a UR5 robot to remotely transport a ferromagnetic particle in a real-world scenario to demonstrate the applicability of the proposed approach.
The experimental results on the UR5 robot show an average success rate of 98.86\% over 30 episodes for randomly generated trajectories, demonstrating the viability of the proposed approach for real-life applications. 
In addition, two classical path finding approaches, Attractor Dynamics and the execution extended Rapidly-Exploring Random Trees (ERRT), are also investigated and compared to the RL-based method.	
The proposed RL-based algorithm is shown to achieve performance comparable to that of the tested classical path planners whilst being more robust  to deploy in dynamical environments.

\end{abstract}

\def\abstractname{Note to Practitioners}
\begin{abstract}
Deep reinforcement learning methods have been widely applied in computer games and simulations. However, employing these algorithms for practical, real-world applications such as robotics becomes challenging due to the difficulty of obtaining training samples. This paper predominantly focuses on bridging the gap between simulations and the real-world implementation of a reinforcement learning algorithm for a robotic application in the context of miniaturized drug delivery robots and robotic capsule endoscopes. This paper presents the derivation and experimental validation of a reinforcement learning-based algorithm for controlling a magnetically-actuated small-scale robot within a simplified model of the large intestine in the presence of disturbances. We demonstrate the possibility of training a high-fidelity reinforcement learning algorithm fully within a simulated environment before deploying it as-is in a real-world scenario by carrying out different experiments and simulations. Implementing the presented control framework complements a large body of this work, and the results offer a feasibility study of using reinforcement learning algorithms in practice.
\end{abstract}


\section{Introduction}
\IEEEPARstart{E}{fficient} drug delivery is a difficult task in the medical industry~\cite{lu2018bioinspired}, and traditional approaches such as pills and intravenous therapy have been the primary method of drug delivery for decades \cite{tiwari2012drug,wen2015drug}. One drawback of these traditional administration methods is that a drug cannot be transported directly to targeted tissues/organs to produce a marked effect. On the other hand, carrier-based drug delivery systems (e.g., small-scale delivery robots) could be better suited to accomplish such precisely targeted delivery task \cite{hu2020micro}.

In recent years, there have been substantial efforts to develop soft miniaturized robots capable of navigating the human body and delivering drugs directly to a tumor or other disease sites precisely and non-invasively.

One of the barriers to small-scale drugs moving toward diseased tissues in the bloodstream is the blood vessel's lining\cite{bourzac2012nanotechnology,schuerle2019synthetic}. Therefore, manual delivery processes might still not be as effective as it needs to be\cite{hu2020micro,sun2017closed}. To overcome this barrier, a promising approach is to actively manipulate drugs or swarms of drugs to reach disease sites and push the particles out of the bloodstream and into target sites \cite{schuerle2019synthetic}. One potential solution to address this issue and increase the efficiency of the delivery process is to robotize the process combined with a remote magnetic actuation mechanism\cite{sun2017closed}. This allows surgeons to have better control over the navigation of drugs even when faced with biophysical barriers, which would otherwise prevent the circulation of drugs.

 The miniaturization of conventional robots is limited mainly by the mechanical structures required to allow them to navigate. By contrast, magnetic actuation is a mechanism that can remotely and wirelessly actuate robots without necessitating specialized structures directly in the robots, thus substantially reducing their size and complexity. Reducing the size of robots enables them to move to difficult-to-reach areas inside biological bodies by applying an external magnetic field that harmlessly penetrates patients' body and applies wrenches on the robots.

Magnetic manipulation of micro/milliscale robots with an application such as magnetic drug delivery has been an increasingly popular approach where drug particles in the body are carried and manipulated by external magnetic fields to reach a targeted location \mbox{\cite{HOSHIAR2017181,hoshiar2018novel,amin2017osmotin,hoshiar2017studies}}. Moreover, small-scale robots/particles  can essentially be seen as magnetic actuation points for continuum manipulators \cite{tariverdi2020dynamic,tariverdi2021recurrent}. In other words, devising control and path planning algorithms for micro/milliscale robots  can eventually contribute to making magnetic continuum manipulators operational in a (semi-)autonomous mode\cite{tariverdi2022physics}.

Remote and robotic-assisted particle or micro/milli-robots manipulation by external magnetic fields has been investigated through traditional control methods, which are reviewed. The work in \cite{khamesee2002design} proposes PID controls and an adaptive control law to remotely operate a microrobot with 3 DOF in an enclosed environment. The designing of classical proportional controllers for controlling two micro-scale magnetic particles in two dimensions while considering interaction wrenches between the particles has been investigated in \mbox{\cite{salehizadeh2017two}}. Driller et al. in \mbox{\cite{diller2013independent}} propose a vision-based proportional feedback controller for motion control and path following a set of two microrobots in three dimensions using magnetic gradient pulling. {In the work \mbox{\cite{khamesee2002design,salehizadeh2017two,diller2013independent}}, particles move in an obstacle and disturbance-free space, and the proposed controllers are not capable of dealing with uncertainties in the case of considering a complex environment.} Zarrouk et al. \mbox{\cite{zarrouk2019four}} address an open-loop control of a particle by using an array of permanent magnets mounted on a robotic arm end-effector. {This work does not consider a localization system to recover the actual position of the particle, and therefore, the proposed open-loop method cannot efficiently accomplish a task such as drug delivery within a noisy environment like the human body.} The previous work \mbox{\cite{ryan2017magnetic}} used a configuration of three rotating permanent magnets and employed a vision-based proportional control strategy in which gain coefficients are tuned by multiple trials. {Although a vision-based localization sensor is used and the magnitude and direction of the applied magnetic field are controllable in this work, it lacks the ability to reject disturbances. In addition, if an obstacle blocks the particle, the considered task cannot be accomplished.} Keller et al. in \mbox{\cite{keller2012method}} propose a human-in-the-loop navigation method to control positions and orientations of a magnetically guided capsule endoscopy. {The presented method in \mbox{\cite{keller2012method}} includes a human in the control loop to manually navigate a capsule endoscope for stomach examinations with multiple functionalities tilting and jumping etc.} The work in \mbox{\cite{diller2011control}} investigates two open-loop control methods for positioning two and three microrobots in two-dimensional space. The microrobots are geometrically designed to respond uniquely to the same applied magnetic fields. Therefore, individuals and subgroups of microrobots can locomote to desired positions by controlling the magnetic field. {The method in \mbox{\cite{diller2011control}} is applied to micro-scale particles (all dimensions under 1 mm), and the global positioning of particles is done in 2-D space using an offline parameter fitting method. } Pieters et al. in  \mbox{\cite{pieters2015rodbot}} propose a model-free vision-based kinematic controller for a microrobot constrained to non-holonomic motions. Kim et al. \mbox{\cite{kim2014closed}} considers a vision-based closed-loop control of an array of magnets on an arbitrary predefined path in three dimensions. The authors in \mbox{\cite{xu2015planar}} propose and compare both open and closed-loop vision-based controllers for steering a helical microswimmer. {The approaches described in \mbox{\cite{diller2011control,pieters2015rodbot,kim2014closed,xu2015planar}} are not equipped to cope with undersecretaries (obstacles and perturbations) within the workspace.} {Although the modern optimal and reinforcement learning control methods for robots have attracted particular attention of scholars in the field of robotics in recent years \mbox{\cite{wang2019unified,peng2019symplectic,jiang2021coordinated}}, real-world implementation of high-fidelity optimal control and RL strategies have not been the point of focus.} 

{An RL-based model-free control of drug dosing for cancer treatment is presented in \mbox{\cite{padmanabhan2017reinforcement}}. The authors use a Q-learning method to schedule the dosing to maximize immune cells for patients. Nevertheless, the presented method is not categorized among targeted drug delivery approaches. In the paper \mbox{\cite{bejar2018deep}}, a Q-learning method with the actor-critic architecture and the deep deterministic policy gradient algorithm is investigated to navigate a particle within an undisturbed 2-D space. The approach uses a linearized model of the system around an equilibrium point to prepare the necessary feedback signals for the controller. However, the approach is not tested in the presence of perturbations. Xu et al. in \mbox{\cite{xu2021reinforcement}} proposes an RL control of a flexible magnetically actuated joint. The study focuses primarily on the tilt motion control for the tip of the instrument, rather than the navigation of the entire joint, as the primary objective.}

{Reinforcement learning algorithms have also been used in a variety of applications, including collaborative assembly tasks \mbox{\cite{zhang2022reinforcement}}, space capture missions \mbox{\cite{jiang2022integrated}}, and 3D navigation in cluttered environments \mbox{\cite{hu2021sim}}.
Zhang et al. in \mbox{\cite{zhang2022reinforcement}} employ a deep deterministic policy gradient algorithm for the task allocation problem in human-robot collaborative assembly tasks. Jiang et al. in \mbox{\cite{jiang2022integrated}} also modify a deep deterministic policy gradient algorithm to develop an adaptive tracking controller based on a highly nonlinear dynamic model of a three-module cable-driven continuum robot. Furthermore, Hu et al. in \mbox{\cite{hu2021sim}} propose a reinforcement learning-based point-to-point navigation algorithm for a mobile robot in a 3D constrained terrain environment.}

By investigating the existing results on magnetic control of small-scale robots (as were reviewed earlier), the following issues are noteworthy: The existing studies mainly employ open-loop or closed-loop control mechanisms that are incapable of accomplishing a given task in uncertain environments. In other words, in practice, the methods are not robust in the presence of disturbances. In addition, the existing studies do not take into account any obstacles in the design of controllers or trajectory planners, so these methods are not suitable for realistic scenarios. 

{It is worth mentioning that coping with disturbances and uncertain environments, together with tracking trajectories or set-point regulations, are the primary factors in controller design and trajectory planners. An interesting point in the magnetic control of milliscale robots is that the mechanism allows teleoperating particles wirelessly. Usually, remote control mechanisms are used when working in an environment that is hazardous or not feasible due to confined or difficult-to-reach areas. This makes the control of magnetic particles more prone to uncertainties.}

{ The general motivation behind the presented method is bridging the gap between simulations and the real-world implementation of a reinforcement learning algorithm for a robotic application. When it comes to obstacle avoidance and perturbation handling scenarios, many approaches have been proposed to solve the challenges. Some methods such as \mbox{\cite{lumelsky1990incorporating,borenstein1991vector,simmons1996curvature}}  achieve local optima and therefore can not guarantee a feasible obstacle-free path. Some others \mbox{\cite{diankov2007randomized,burns2005toward,sprunk2011online}} do global searches for a feasible, valid path, yet the heavy computational costs prevent them from reacting properly to pop-up obstacles.
	
RL-based methods can offer a hybrid behavior by learning a dynamic, time-varying environment. These approaches do re-planning and path deformation on-the-fly to avoid suddenly appearing obstacles and find a valid global path to a determined target. In addition, compared to those methods in which global explorations happen online, the training of RL methods is offline; hence, they are reactive during the execution of a task. In other words, if the deformation of a path does not result in a valid trajectory toward a target, a new valid path is calculated. It is worth mentioning that the quality of the behavior is highly dependent on the samples used for offline training. Also, the system's performance can only be guaranteed statistically. At the same time, it is feasible to generalize and implement the approach for different environments with varying forms of obstacles and perturbations, provided that more samples of environments are used for offline training and proper tweaking of the controller's hyperparameters is considered.}

{To evaluate results of the RL-based mechanism, two classical approaches, Attractor Dynamics\mbox{\cite{khansari2012dynamical}} and execution extended Rapidly-exploring Random Trees (ERRT)\mbox{\cite{bruce2002real}} are tested through conducting simulations and experiments in which the results confirm that the RL-based method achieves comparable performance while being a more robust and generic solution to deploy in dynamical, complex environments.}

Based on the abovementioned issues, more practical approaches to magnetic control of small-scale robots require investigation. Although designing online path planners with appropriate robust controllers is a problem-specific approach, deep Reinforcement Learning (RL) methods can handle both the trajectory planning and control tasks simultaneously and can be generalized to be used in various workspaces. This paper proposes to employ a deep RL algorithm for magnetic position control of an object in a constrained environment.
{To the best of the authors' knowledge, this is the first time that magnetic control of milli-sized (ferro)magnetic particles using an RL algorithm capable of avoiding obstacles and rejecting disturbances is addressed.} Therefore, the main contribution of this work are as follows.
 \begin{itemize}
 	\item   {Implementing an RL-based switching framework for dynamic control of a particle in the presence of obstacles in two dimensions with visual feedback and using a robotic arm in a real-world scenario.}
 	
 	\item {Through simulations and experiments, the results of the proposed RL-based algorithm and classical approaches such as Attractor Dynamics, and execution extended Rapidly-exploring Random Trees (ERRT) are compared,
 	revealing the practical efficacy of the proposed
 	methodology.}
 \end{itemize}

This paper is organized as follows: The problem statement is given in Section II. Preliminaries on the RL and Rainbow algorithms are provided in Section III. {Section IV thoroughly describes our proposed RL-based, Attractor dynamics-based, and ERRT-based switching controllers. Section V is devoted to explaining the simulation and experiment setups, and the results from the simulations and real-world scenarios, while their associated discussion is presented in Section VI.}

\section{Problem Statement}
Inspired by the literature on small-scale robot control and to address more practical problems (e.g., scenarios in the presence of obstacles and disturbances), this paper aims to propose and implement an RL-based switching framework for dynamic control of a particle in a constrained environment using a robotic arm and visual feedback. In this work, we have employed an RL algorithm in a switching control framework such that the first sub-controller is responsible for finding and tracking the magnetic particle through a camera mounted on the end-effector of a robotic arm. This sub-controller makes the whole system robust against disturbances. A second sub-controller makes the robot arm carry the particle through a constrained environment toward a target position. In addition, a simplified model of the large intestine in two dimensions has been employed as a constrained environment in which a milliscale robot is being carried remotely using an external magnetic field generated by a permanent magnet mounted on a robotic arm.

The strength and direction of a magnetic field produced by an attached permanent magnet on the end-effector are not considered to be adjustable, and due to interference introduced by the environment, a magnetic particle may not respond to the field and be carried within the environment. To guarantee that the particle responds to the external field, we assume that an external magnetic field is effective on a magnetic particle, i.e., a milliscale robot, provided that the particle is within a specific distance from the end-effector of the robotic arm.

 {As disturbances, we physically draw the magnetic particle away from the magnetic control zone during active control within the constrained environment ---a simplified model of the large intestine---. In other words, disturbances are displacements in 2-D space. We assume that disturbances do not occur too quickly and/or with large accelerations. It is worth mentioning that in real-world scenarios such as the drug delivery application, disturbances such as movement of the GI tract or a fluid current running into a vessel or the tract might hold the particle back from a region where the wrenches of the end-effector's magnet are effective.}

\section{Preliminaries}

RL deals with learning the behavior (policy) of an agent, which in this work is a robot that interacts with its environment to perform a task. It is an approach for solving control problems in which a control policy is learned through repeated trial-and-error interactions between the robot and its environment. In RL, the behavior of the robot is learned to maximize the expected sum of rewards provided by its environment through feedback signals.  Assuming discrete time steps $t$, the robot performs an action $a_t \in \mathds{R}^n$, $n\in\{1,2,\cdots\}$ that depends on the current state of the system $s_t \in \mathds{R}^n$ and is also considered as a control action. This action result in a new state $s_{t+1}$, and the robot receives a reward $r_{t+1}$. This process is repeated, and the robot's goal is to learn the optimal policy $\pi$, which maximizes the expected cumulative rewards.

The traditional Q-Learning \cite{watkins1992q} is concerned with learning Q-values. Q-values are the expected accumulated rewards the robot receives when following a given policy starting from the state-action pair $(s_t, a_t)$. In the traditional Q-learning algorithm, we construct a Q-value table containing all the Q-values mapping between the possible states a robot can get by moving between the different states. This basic algorithm can be inefficient in sampling the state space and may get stuck in a local optima \cite{mnih2013playing}.
Deep Q-Networks (DQN) algorithm has been an important milestone introduced by \cite{mnih2013playing}. The difference between DQN and its table version is that instead of having all the Q-values stored in a look-up table, they are represented as a multi-layered neural network such that, for a given state $s$, outputs a vector of action values $Q(s,\theta)$, where $\theta$ are the parameters of the network. Also, by approximating these Q-values via a neural network, one can consider action and state space that could not realistically be fitted in memory (mainly for embedded applications). This results in more significant generalization attributes and richer representation.

Despite its usefulness, several limitations of the DQN algorithm are now known, and many extensions have been proposed so far to enhance its speed or stability. Some limitations are as follows: Because of a max operator in the Q-learning equation, DQN may overestimate Q-values and then choose sub-optimal actions in some states. In addition, uniformly sampling from a replay memory in the experience replay mechanism is not optimal \cite{van2015deep}. In other words, prioritizing samples in a replay memory based on some criteria, such as the error of the estimated Q-values and the actual Q-values, can help mitigate the estimation deviations \cite{schaul2015prioritized}. Furthermore, by introducing a function (i.e., the advantage function) to DQN, comparing the goodness of actions can be done \cite{wang2016dueling}. Moreover, unlike the DQN algorithm that finds out about the mean or max of Q-values, learning their distributions has shown a better performance \cite{bellemare2017distributional}. In noisy networks, weight parameters are modeled as distributions. In other words, weight parameters are sampled from some distributions. However, in noisy networks, predicted actions are covered by distributions, and then finding the best action is more challenging which can be done by reducing the size of sampling
distributions or data uncertainty over time \cite{fortunato2017noisy}. The deep reinforcement learning community has made gradually several improvements to the DQN algorithm \cite{mnih2013playing} to cope with its shortages, for example, Double Deep Q-learning Networks  (DDQN) \cite{van2015deep}, Prioritized DQN \cite{schaul2015prioritized}, Dueling DQN \cite{wang2016dueling}, Distributional DQN \cite{bellemare2017distributional}, Noisy DQN \cite{fortunato2017noisy}. Finally, Rainbow \cite{hessel2017rainbow} combines the improvements as mentioned above.

\begin{figure*}[htbp]
	\centering
	\includegraphics[width=\textwidth]{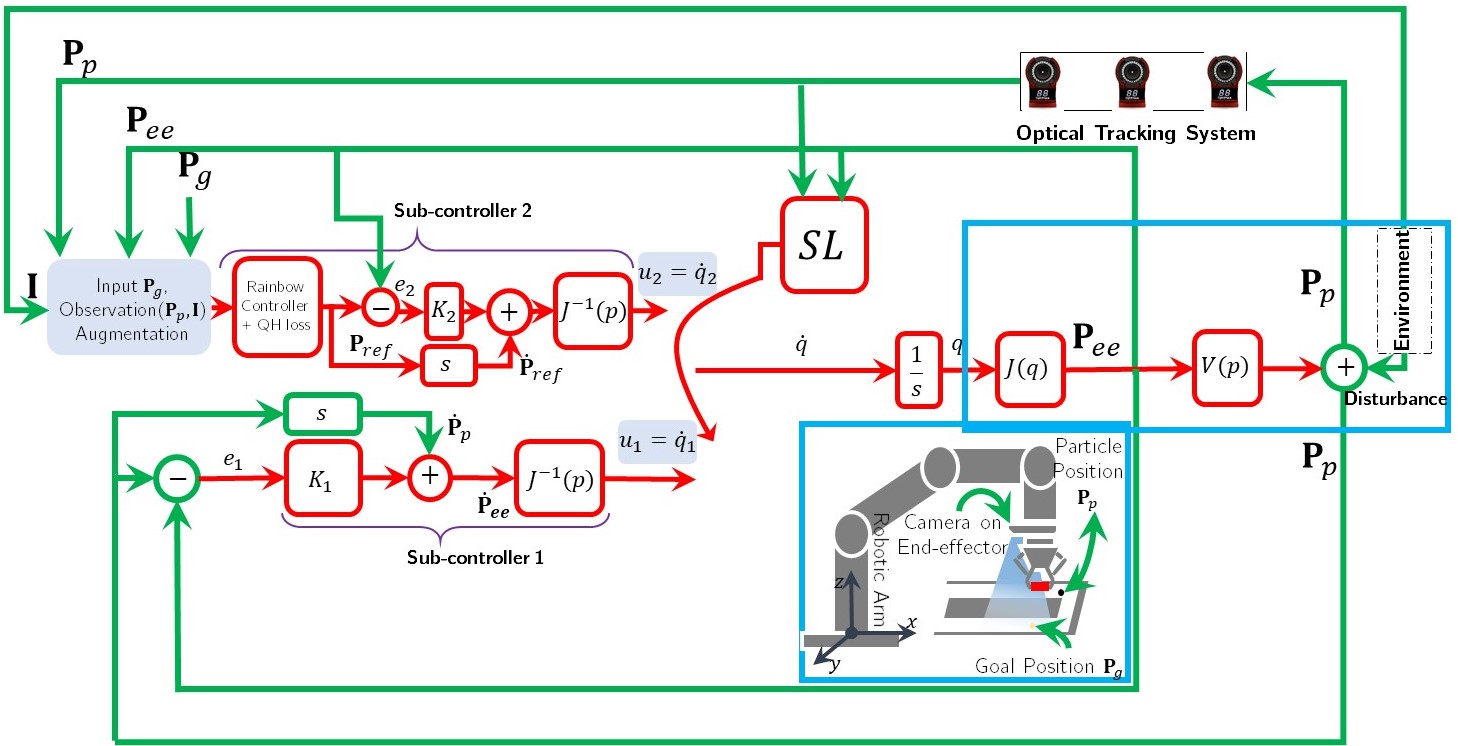}

	\caption{ \protect{The switching controller architecture consists of Sub-controllers 1 and 2 and a switching law ($ SL$). $ SL$ is a supervisory algorithm that controls the switching between the various controllers. The observations or feedback signals are: the magnetic particle position obtained through the optical tracking system ($ \mathbf{P}_p$), partial information ($ \mathbf{I}$) of the constrained environment captured by an RGB camera on the end-effector, i.e., the RGB camera only covers a portion of the environment for the Sub-controller 2. $ s $ represents a derivative filter. In Sub-controller 1, $ K_1 $ represents a proportional gain for adjusting the asymptotic convergence of the controller, and in Sub-controller 2, $ K_2 $ represents a positive definite 3-by-3 matrix for adjusting the asymptotic convergence of the controller. $ K_2 $ is a simple model for the internal control of the robot.   Joints are modeled as a single integrator which is represented by $ \frac{1}{s} $. Inverse kinematic and kinematic of the robot is denoted by $ J^{-1}(p) $ and $ J(q) $, respectively. $ q $ and $ \dot{q} $ represent joint positions and angular velocities. $ V(p) $ models the unknown magnetic interaction dynamic between the external magnet on the end-effector and the particle with the input $ \mathbf{P}_{ee}$ and output $ \mathbf{P}_p$.   }}
	\label{fig:cascadeController}
\end{figure*}

 The main characteristics of the Rainbow algorithm are as follows:  an experience replay memory is introduced, in which experiences are stored consecutively. Uniformly random sampling from this buffer removes correlation between the observations to some extent. In the employed Rainbow algorithm, an $N$-step replay buffer is implemented where the most recent $N$ observations are stored. The $N$-step replay buffer is used in the $N$-step learning method \cite{sutton1988learning}. Prioritized replay buffer \cite{schaul2015prioritized} is another crucial feature used in Rainbow, where a criterion is used to measure how unexpected a transition is, and based on this, the most unexpected transition is replayed from memory, and weight and bias parameters are updated accordingly. Noisy networks are an alternative for the $\epsilon$-greedy method used in DQN. In noisy layers, deterministic weight and bias parameters are replaced by a unit Gaussian distribution, and the applied noises to weight and bias parameters are independent. In Dueling architecture \cite{wang2016dueling}, there is a possibility to compare the goodness of all actions in a specific state by introducing advantage and value layers. Also, in \cite{bellemare2017distributional}, the authors discuss a method known as Categorical DQN, which learns the distribution of returns instead of the expected return, and using the proposed method, returns distributions satisfy a variant of Bellman’s equation. In the implemented Rainbow, a Dueling architecture is integrated with the Categorical DQN method, and the last two layers of advantage and value are replaced by the noisy layers.

\section{Methods: Reinforcement Learning, Attractor Dynamics, and ERRT-based Switching Controllers}

This section primarily discusses the proposed control architecture in which
a switching control architecture comprises two controllers, i.e., Sub-controllers 1 and 2, and a switching law. Sub-controller 1 consists of the inverse kinematic of the robot and keeps the magnetic particle within the immediate vicinity of the robotic arm end-effector. Once the particle is far enough from the end-effector, Sub-controller 1 takes the robotic arm to the particle position, which is feedbacked into Sub-controller 1 from an optical tracking system. As soon as the particle is again close enough to the end-effector, an appropriate switching law makes Sub-controller 2  control the robot arm to carry the particle toward an objective inside the environment using a magnetic field produced by an attached permanent magnet on the end-effector.

\subsection{Reinforcement Learning-based Approach}

In the RL-based method, Sub-controller 2 is trained beforehand both as a trajectory planner to prevent the robot from crashing into obstacles and also as a controller, which makes the robot globally stable by generating trajectories that always carry the particle toward a
target for an arbitrary initial position. Sub-controllers 1 and 2 and the switching law are shown in Figure~\ref{fig:cascadeController}.

\subsubsection{Sub-controller 1}\label{RLMethod}

\begin{figure}
    \centering
	\includegraphics[width=0.45\textwidth]{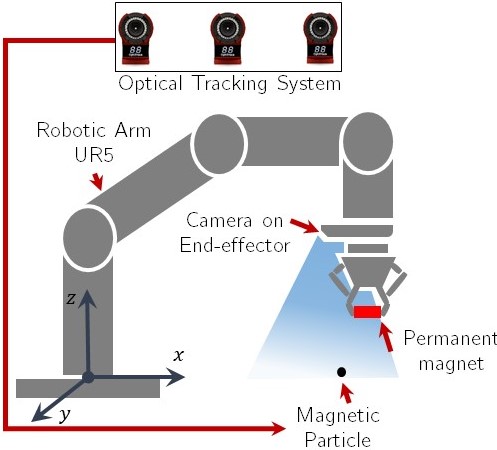}
    \caption{{The control scheme employs analytical inverse kinematic solutions when Sub-controller 1 is activated: The feedback coming from the optical tracking system consists of the particle position, and by employing the analytical inverse kinematic solutions, the robotic arm will locate the particle.}}
    \label{fig:innerloop}
\end{figure}

Since there is no control on the strength and direction of a magnetic field produced by an attached permanent magnet on the end-effector and due to uncertainties and disturbances induced by the environment, the control system should be robust to losing the particle (when the particle is not responding to the field) and able to be automatically retrieved if such an event occurs. Sub-controller 1 is responsible for keeping the magnetic particle within the vicinity of the end-effector through the information it gets from an optical tracking system. In other words, the robotic arm should be capable of finding the magnetic particle or tracking the particle starting from an arbitrary initial position. When Sub-controller 1 is activated, the control-loop mechanism is depicted in Figure~\ref{fig:innerloop}.

{IKFast, The Robot Kinematics Compiler (\mbox{\cite{diankov2010automated,zeng2018learning}}), which analytically solves the inverse kinematics equations of the robotic arm, UR5 in this work, is employed as Sub-controller 1 to map Cartesian space of the end-effector to the joint space of the UR5.} {Furthermore, the locus of the end-effector remains parallel to the plane where the particle is manipulated, i.e., the angle between the plane of work and the z-axis should remain unchanged.} 

\begin{figure}
	\centering
	\includegraphics[width=0.45\textwidth]{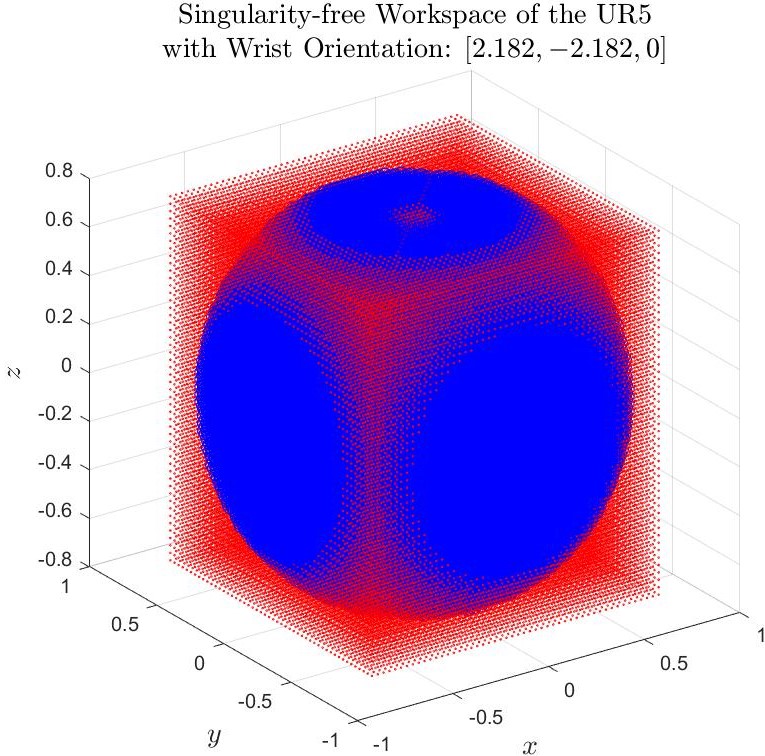}
	\caption{\protect{The blue area shows the singularity-free workspace of the UR5 with the wrist orientation: $[2.182, -2.182, 0]$. Considering the locus of the end-effector when UR5 manipulates the particle within the working environment, the manipulator will not encounter a singularity (red-colored volume).}}
	\label{fig:singlulFree}
\end{figure}

{When a disturbance occurs, Sub-controller 1 will be activated and the robot basically employs a kinematic control. Therefore, performance can be very satisfactory, provided that inserted disturbances are not too fast and/or do not require large accelerations. Also it should be noted that one of the issues for using a kinematic control method is singularities. This implies that the robotic arm needs to be singularity-free within the workspace where the simplified structure of the large intestine is located. Outside of the workspace, the RL algorithm will assure that the arm does not get stretched too beyond the region of interest. The singularity analysis of the employed robotic arm (UR5) should be taken into consideration. As the locus of the end-effector remains parallel to the plane where the particle is manipulated, Figure~{\ref{fig:singlulFree}} shows that the arm will not encounter singularities.}

{In the presented scenario, switching between different sub-controllers essentially implies altering desired joints angles ---calculated from desired values in Cartesian space--- and a closed control architecture within industrial robots or cobots (collaborative robots) such as UR5 has the responsibility of regulating joint angles to desired values (sometimes for simplicity each joint of an industrial arm might be seen as a simple integrator, which is being controlled internally). It should be noted that the due to the nature of tasks considered in this work which are set-points regulations ---kinematics-based when Sub-controller 1 is active, or based on simplified dynamics (single integrators) when Sub-controller 2 is active---, disturbances do not occur too fast and/or with large accelerations. }

\subsubsection{Sub-controller 2}
\begin{figure}
    \centering
	\includegraphics[width=0.45\textwidth]{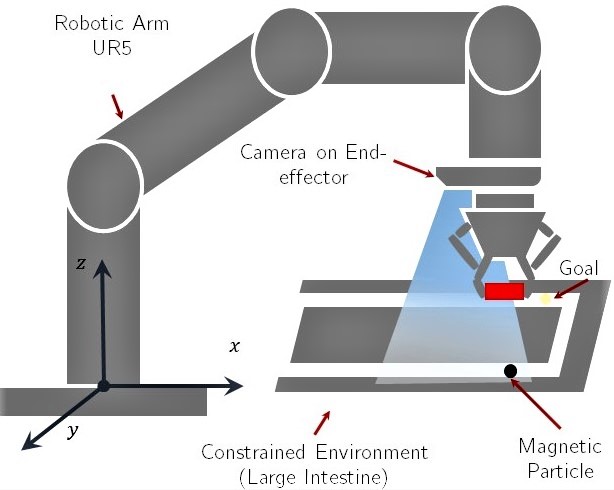}
    \caption{{The control scheme that trains a customized Rainbow algorithm to navigate through a constrained workspace when Sub-controller 2 is activated: Sub-controller 2 steers the robotic arm to carry the particle to a desired location in the environment.}}
    \label{fig:outerloop}
\end{figure}

Once the robotic arm locates the magnetic particle, Sub-controller 2 steers the robotic arm to carry the particle to a desired location in a constrained (intestine-like) environment. Importantly, feedback of the optical tracking system is not used in this controller ($ \mathbf{P}_{ee} $ is in the vicinity if $ \mathbf{P}_{p} $), and the robots only use the end-effector RGB camera to perceive the environment partially.  When Sub-controller 2 is activated, the control-loop mechanism ---depicted in Figure~{\ref{fig:outerloop}}--- is as follows: The RL algorithm takes the particle toward a user-determined target position within the environment. This sub-controller is active as long as the particle is responsive to the external magnetic field of the magnet attached to the end-effector, in other words, the particle is within a specific threshold from the end-effector.

\newsavebox{\smlmat}% Box to store smallmatrix content
\savebox{\smlmat}{$ \begin{smallmatrix}\mathbf{G}_{2 \times 2} = [K~{Id}]\end{smallmatrix} \left[\begin{smallmatrix}\mathbf{A}&\mathbf{B}\\{Id}&\mathbf{0}\end{smallmatrix}\right]^{-1} [\begin{smallmatrix} \mathbf{0}\\{Id}\end{smallmatrix}]$}

{For Sub-controller 2, inspired by \mbox{\cite{hessel2017rainbow}}, a customized Rainbow algorithm with Quantile Huber loss from Implicit Quantile Networks (IQN) algorithm\mbox{\cite{dabney2018implicit}} and ResNet\mbox{\cite{he2016deep}} are employed. The Huber loss is traditionally used to train the family of DQN RL algorithms since the introduction of DQN  by Mnih et al. in \mbox{\cite{mnih2015human}}. A comparative study by Obando-Ceron et al. in \mbox{\cite{obando2020revisiting}} showed the superior performance of the Huber loss over the more straightforward mean-squared error loss.}

\begin{figure}
	\centering
	\includegraphics[width=0.5\textwidth]{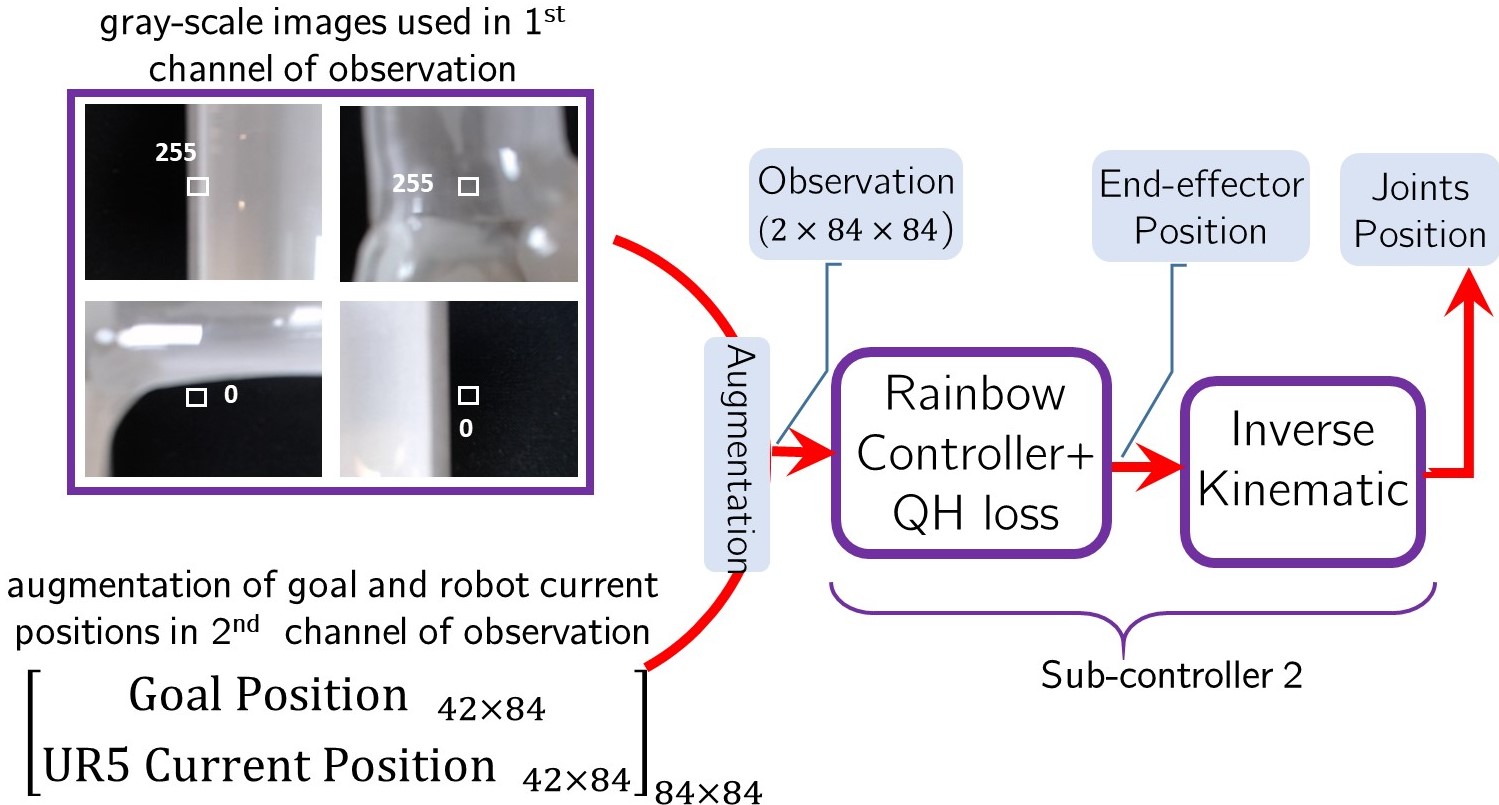}
	\caption{Observations to train Sub-controller 2: Two-channel images $84 \times 84$, i.e., $2 \times 84 \times 84$. The first channel includes a gray-scale image obtained from the end-effector RGB camera, and the second channel is an augmentation of two matrices with a size of $42 \times 84$ which are populated with the goal and robot current positions. The average of the pixel values in the center of each gray-scale image is used for obstacle detection.}
	\label{fig:train}
\end{figure}

\begin{figure*}
	\centering
\includegraphics[width=\textwidth]{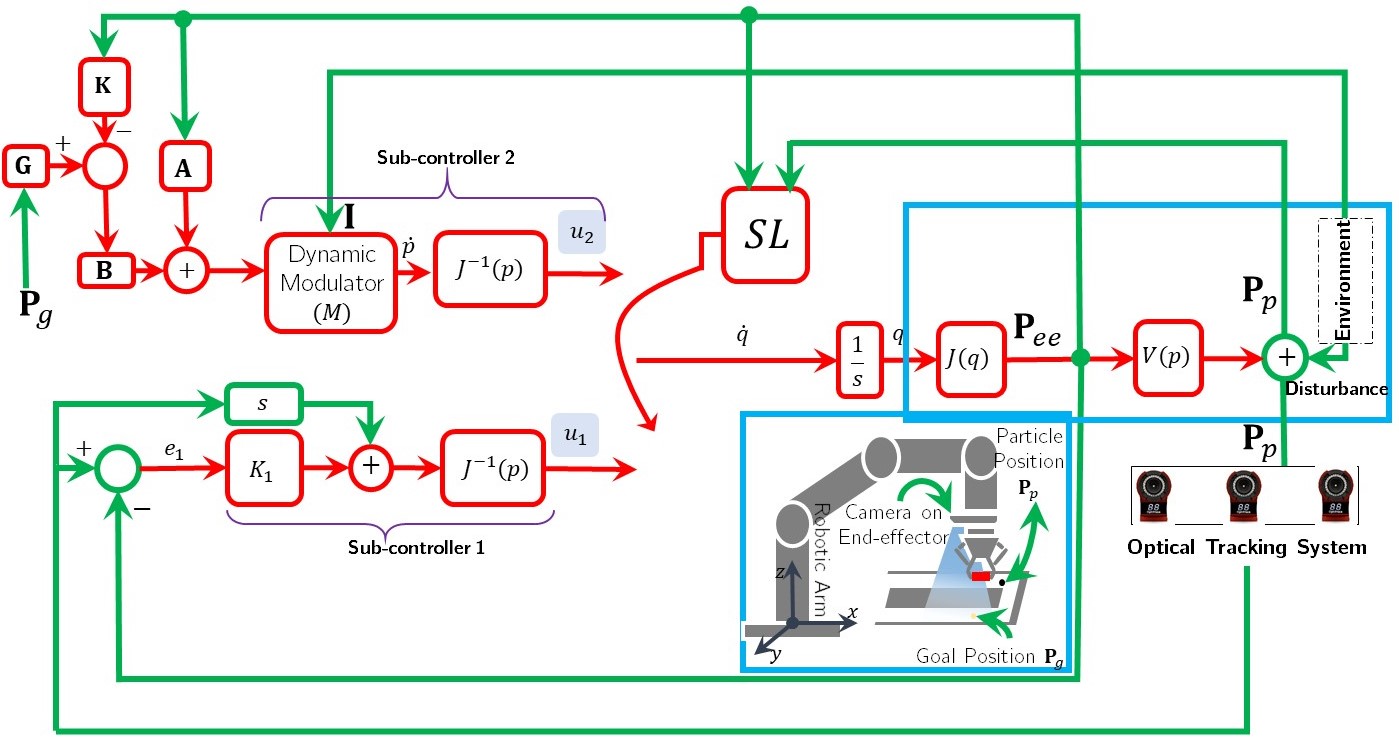}	
	\caption{The switching controller architecture consists of Sub-controllers 1 and 2 and a switching law ($ SL$). $ SL$ is a supervisory algorithm that controls the switching between the various controllers.
	The observations or feedback signals are: the magnetic particle position obtained through the optical tracking system ($ \mathbf{P}_p$), partial information ($ \mathbf{I}$) of the constrained environment captured by an RGB camera  on the end-effector, i.e., the RGB camera only covers a portion of the environment for the Sub-controller 2. $ \mathbf{K}$ is the gain of the state feedback controller,  $ \mathbf{B}_{2 \times 2}$, and $ \mathbf{A}_{2 \times 2}$ are the identity matrices ($ {Id}_{2\times 2} $). Furthermore, ~\usebox{\smlmat}. The symbol $ s $ represents a derivative filter. In Sub-controller 1, $ K_1 $ represents a proportional gain for adjusting the asymptotic convergence of the controller. Joints are modeled as a single integrator which is represented by $ \frac{1}{s} $. Inverse kinematic and kinematic of the robot is denoted by $ J^{-1}(p) $ and $ J(q) $, respectively. $ q $ and $ \dot{q} $ represent joint positions and angular velocities. $ V(p) $ models the unknown magnetic interaction dynamic between the external magnet on the end-effector and the particle with the  input $ \mathbf{P}_{ee}$ and output $ \mathbf{P}_p$.}
	\label{fig:attractorDyn}
\end{figure*}

Observations which are fed to Sub-controller 2 are two-channels images with dimension $84 \times 84$, i.e., $2 \times 84 \times 84$. It is worth mentioning that images are downsampled from $480 \times 480$ to $84 \times 84$. The first channel includes a gray-scale image obtained from the end-effector RGB camera, and the second channel is built by augmentation of two matrices with the size $42 \times 84$, which are populated with the goal and current positions of the robot, respectively (Figure~\ref{fig:train}). It should be noted that a simple segmentation process is done on each RGB camera to detect obstacles. A square of pixels of size 15-by-15 in the center of each image is considered, and then the mean of the pixel values in that square is calculated and used to detect obstacles. An obstacle is detected when the pixels' average is $<$150. Four discrete actions are considered for controlling the robotic arm end-effector in the $xy$-plane in two dimensions by setting a new position for the end-effector with a small change with respect to the end-effector current position. Figure~\ref{fig:train} shows details of the training.

Furthermore, as a reward function employed in the training of Sub-controller 2, {an episode is terminated with a negative reward if there is no answer for the inverse kinematic} or number of steps in each episode exceeds 150 or it is detected that the end-effector would force the particle inside an obstacle. In addition, an episode is terminated with a positive reward if a goal is detected by the end-effector-camera within 2~cm in the $xy$-plane. In precise terms, if the Norm-2 of the error between a current position of the robotic arm and a goal position is $<=$ 2~cm, that target is reached. It should be added that during training, goals are spawns randomly at different locations in each episode.

\subsubsection{Switching Signal}
%\hlyellow
{In the switching control architecture, the updating law for activating the sub-controllers is as follows: }

\begin{equation}\label{switchingLaw_v2}
%	SL = \frac{3+sgn(\mathbf{\text{T}} - e)}{2}
	SL = \frac{1}{2} \Big(3+sgn(\mathbf{\text{T}} - e)\Big)
\end{equation}
%\hlyellow
{where $ SL $ is the index of to-be-active Sub-controller, e = $ \|\mathbf{\text{P}}_{ee} - \mathbf{\text{P}}_p\| $ and $ \mathbf{\text{P}}_{ee} $ is the current position of the robotic arm, $ \mathbf{\text{P}}_{p} $ is the position of the particle, and $ \mathbf{\text{T}} $ is an arbitrary threshold set to 10 cm for the experiment. Although $ \mathbf{\text{T}} $ is an arbitrary threshold, it is dependent on the step size of the manipulator, transporting the particle and also the particle's size.  In other words, the reduced threshold is allowed by using the smaller step or particle sizes.

It should be noted that $ sgn(\text{arg}) = 1$, if $\text{arg}>=0 $, otherwise $ sgn(\text{arg}) = -1$.
In other words, Sub-controller 2 is activated unless the euclidean distance between end-effector and particle positions is more than or equal $ \mathbf{\text{T}} $. }

\subsection{Attractor Dynamics-based Approach} \label{AttD}
%\hlyellow
{In the Attractor Dynamics-based method, a state feedback controller together with a dynamic modulator \mbox{\cite[Section 3.2]{khansari2012dynamical}} are designed to ensures that the robot will not hit  convex obstacles while carrying the particle toward a goal. The block diagram of the algorithm is shown in Figure~{\ref{fig:attractorDyn}}. Sub-controller 1 and Switching law are the same as what is discussed in Section {\ref{RLMethod}}. For Sub-controller 2, each joint of the manipulator is modeled as a continuous single integrator:}

\begin{equation}\label{manipulator}
%	\mathcolorbox{lemonchiffon}
{	\begin{split}
	\dot{p}(t) &= \mathbf{A}p(t) + \mathbf{B}u(t),\\
o(t) &= p(t)
	\end{split}}
\end{equation}
%\hlyellow  
{where} ${p(t) = \begin{bmatrix}
	x \\ y
\end{bmatrix}\in \mathbb{R}^{2\times1}} $ {is Cartesian position of the end-effector in a 2-D space ($ xy $-plane),  $ \mathbf{B}_{2 \times 2}$, state matrix and $ \mathbf{A}_{2 \times 2}$, input matrix, are the identity matrices ($ {Id}_{2\times 2} $). Furthermore, $ u(t) \in \mathbb{R}^{2\times1} $ is a 2-by-1 control input.  Let $ \mathbf{\text{P}}_g $ denotes the desired constant 2-by-1 vector for the output $ o(t) $ to track asymptotically.}

{The control goal of Sub-controller 2 is to design a state feedback controller in which $ u(t) $ depends on $ p(t) $ and $ \mathbf{\text{P}}_g $ so that  the
regulation error $ e(t) = \mathbf{\text{P}}_g - o(t) $ goes to zero when $ t\rightarrow \infty $. It can be shown (\mbox{\cite{brogan1991modern,chen2004linear}}) that the state feedback control law:} \begin{equation}\label{ctrlawstate}
	{u(t) = \mathbf{G}\mathbf{\text{P}}_g - \mathbf{K}p(t) }
\end{equation} 
{makes System {\ref{manipulator}} a globally asymptotically stable system, where} $ {\mathbf{G}_{2 \times 2} = [K~{Id}]_{2\times4} \begin{bmatrix}\mathbf{A}&\mathbf{B}\\{Id}&\mathbf{0}\end{bmatrix}^{-1}_{4\times4} \begin{bmatrix} \mathbf{0}\\{Id}\end{bmatrix}_{4\times2}}$.

{Although the control law {\ref{ctrlawstate}} makes the manipulator stable, it cannot  prevent the robot from colliding with obstacles. Therefore, a real-time
obstacle avoidance strategy should  be considered together with the law {\ref{ctrlawstate}}. To design the obstacle avoidance protocol, we follow the same line of ideas as in \mbox{\cite{khansari2012dynamical}}. In this work, we consider rectangular 2-D obstacles based on superellipse curves, where $ \Gamma(p): \| \frac{x - x_o}{a}\|^n + \| \frac{y - y_o}{b}\|^n  = 1$ represents the boundary points of an obstacle with the center point $ [x_o, y_o] $, furthermore $ a $ and $ b $ are called the semi-diameters of the curve. In other words, the curve $ \Gamma(p) $ is confined in the rectangle $ \| x-x_o \|<=a $ and $ \| y-y_o \|<=b $. The similarity of the curve to a rectangle  is adjustable with the parameter $ n $, and with $ n>2  $, the curve looks like a rectangle with rounded corners.

The dynamic modulator matrix $ M $ (designed in \mbox{\cite[Section 3.2]{khansari2012dynamical}}) propagates the
influence of the obstacle on the motion flow with the maximum effect 
at the boundaries of the obstacle, and vanishes for points far from it. Applying the dynamic modulator, on the linear system {\ref{manipulator}} yields:}

\begin{equation}\label{modulManipulator}
{	\dot{p}(t) = M\Big(\mathbf{A}p(t) + \mathbf{B}u(t)\Big)}
\end{equation}
{where $ u(t) $ is given in Equation {\ref{ctrlawstate}}.
Details of design, stability and convergence analysis of the system with a modulation matrix can be found in \mbox{\cite{khansari2012dynamical}}. Here we skip the analysis for brevity.

For the computer implementation of {\ref{modulManipulator}}, a simple way is to replace the derivative by a difference}
\begin{equation*}
{p(t_{k+1}) 	=   M\Big(\mathbf{A}p(t_k) + \mathbf{B} \mathbf{G}\mathbf{\text{P}}_g - \mathbf{K}p(t_k)\Big)h + p(t_{k})h}
\end{equation*}
{where $ t_k  $ is the sampling instant and $ h = t_{k+1} - t_k $ is the sampling period.}

\subsection{{ERRT-based Approach}}
\begin{figure*}[htbp]
	\centering
\includegraphics[width=\textwidth]{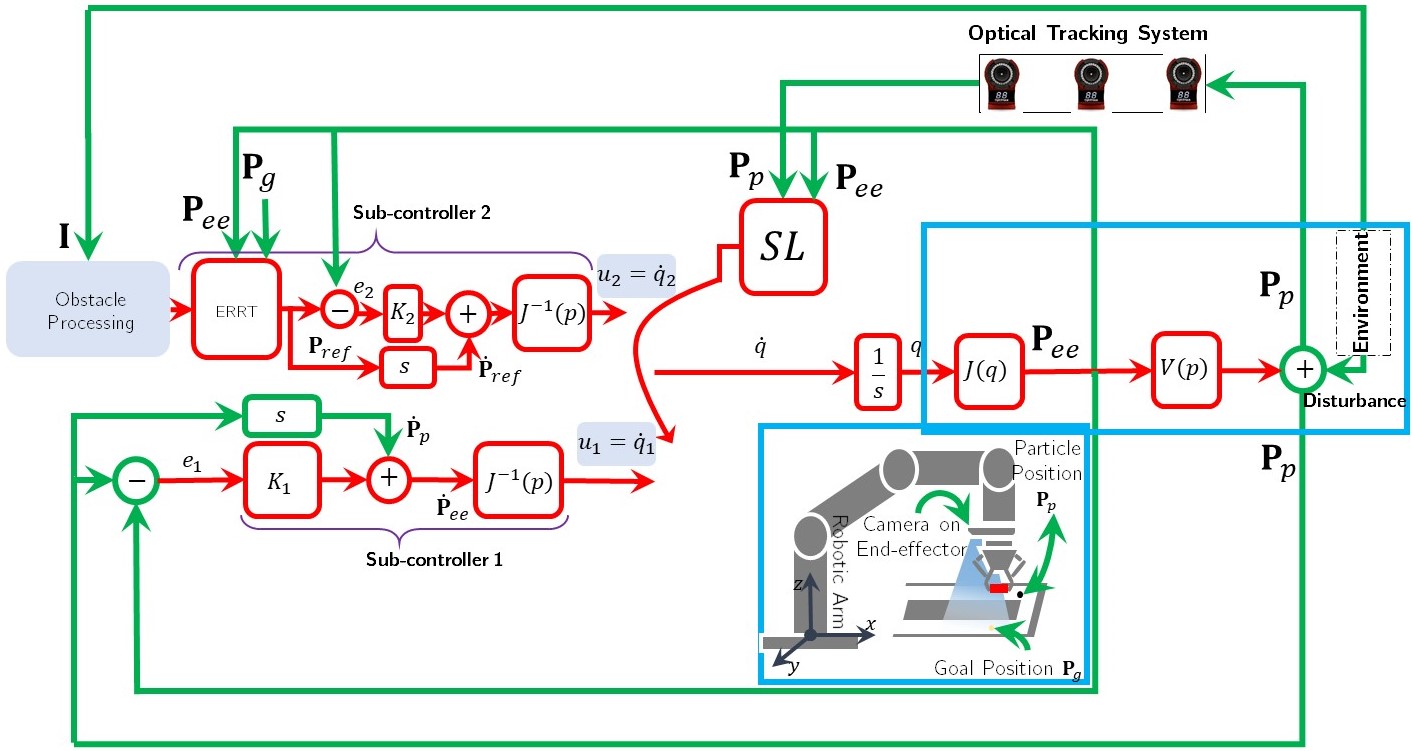}	
	\caption{{The switching controller architecture consists of Sub-controllers 1 and 2 and a switching law ($SL$). $SL$ is a supervisory algorithm that controls the switching between the various controllers. The observations or feedback signals are: the magnetic particle position obtained through the optical tracking system ($ \mathbf{P}_p$), partial information ($ \mathbf{I}$) of the constrained environment captured by an RGB camera on the end-effector, i.e., The RGB camera only covers a portion of the environment for the Sub-controller 2. $ \mathbf{I}$ is fed to the obstacle processing unit which optimally partitions obstacles into minimum number of rectangular obstacles. The ERRT unit generates an obstacle-free path based on the given feedback information. The symbol $ s $ represents a derivative filter. In Sub-controller 1, $ K_1 $ represents a proportional gain for adjusting the asymptotic convergence of the controller, and in Sub-controller 2, $ K_2 $ represents a positive definite 3-by-3 matrix for adjusting the asymptotic convergence of the controller. $ K_2 $ is a simple model for the internal control of the robot. Joints are modeled as a single integrator which is represented by $ \frac{1}{s} $. Inverse kinematic and kinematic of the robot is denoted by $ J^{-1}(p) $ and $ J(q) $, respectively. $ q $ and $ \dot{q} $ represent joint positions and angular velocities. $ V(p) $ models the unknown magnetic interaction dynamic between the external magnet on the end-effector and the particle with the  input $ \mathbf{P}_{ee}$ and output $ \mathbf{P}_p$.}}
	\label{fig:errt}
\end{figure*}

{To better contextualize the proposed RL-based control algorithm, this work  additionally proposes employing an ERRT-based algorithm, the specifics of which is detailed in \mbox{\cite{bruce2002real}} as a path planner within the Sub-controller 2. 
	
ERRT is a sampling-based planning method for a continuous domain.   Its two extensions compared to RRT method \mbox{\cite{lavalle1998rapidly}} ---the waypoint cache and adaptive cost penalty search--- enable the method to explore the environment and plan an obstacle-free path on the fly. However, the quality of the generated path and efficient replanning heavily depends on the processing of the received obstacles images from the environment. Therefore, to be able to use this algorithm to solve the presented problem in this paper, we develop a technique to optimally allocate obstacles in rectangular shapes to be used in the ERRT algorithm. In addition, the applicability of  algorithm in real-time heavily depends on how much the environment is polluted by obstacles, the complexity of obstacle shapes, and also whether obstacles are static or dynamic and how fast obstacles are changing. 

The block diagram of the switching controller based on this approach is presented in Figure~{\ref{fig:errt}}. Sub-controller 1 and Switching law are the same as what was mentioned in Section~{\ref{RLMethod}}. However, in Sub-controller  2, ERRT, instead of the RL-based algorithm, is used as a path planner in Cartesian space and the output is fed to the inverse kinematic function to generate joint angles.}

\begin{figure}
	\centering
	%	\subfloat[a\label{1b}]{T
	\subfloat[\label{obP_a}]{%
		\includegraphics[width=\linewidth]{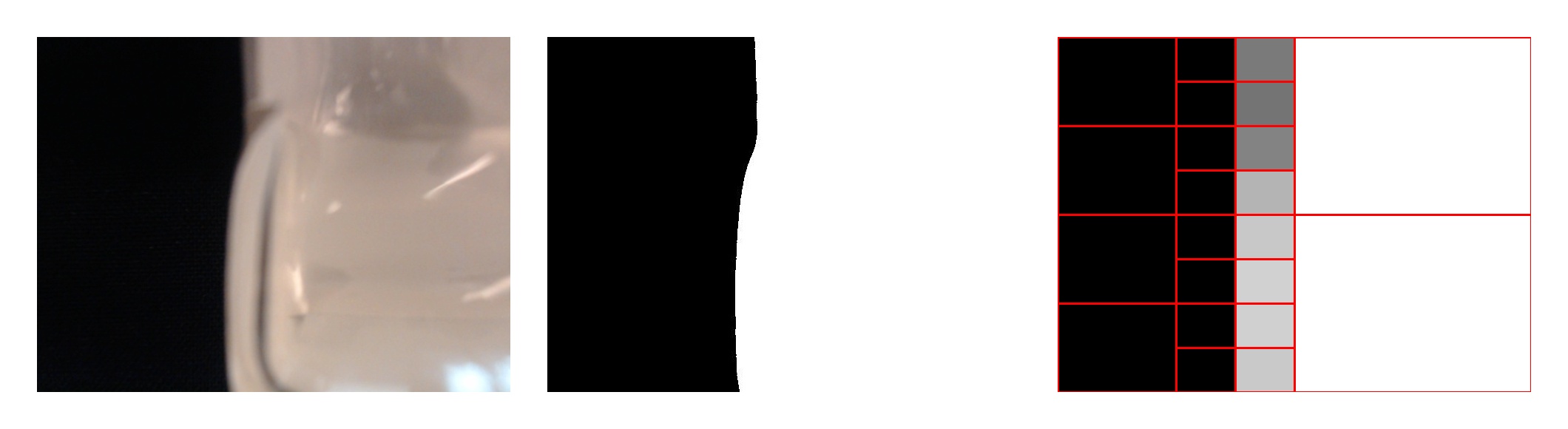}}
	\\
	\subfloat[\label{obP_b}]{%
		\includegraphics[width=\linewidth]{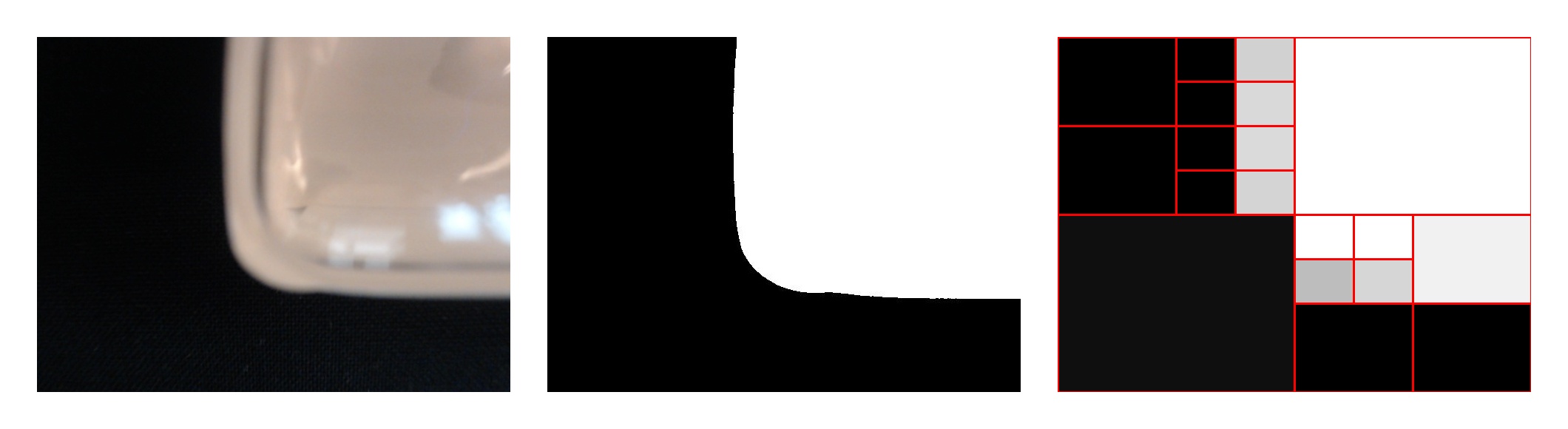}}
	\\
	\subfloat[\label{obP_c}]{%
		\includegraphics[width=\linewidth]{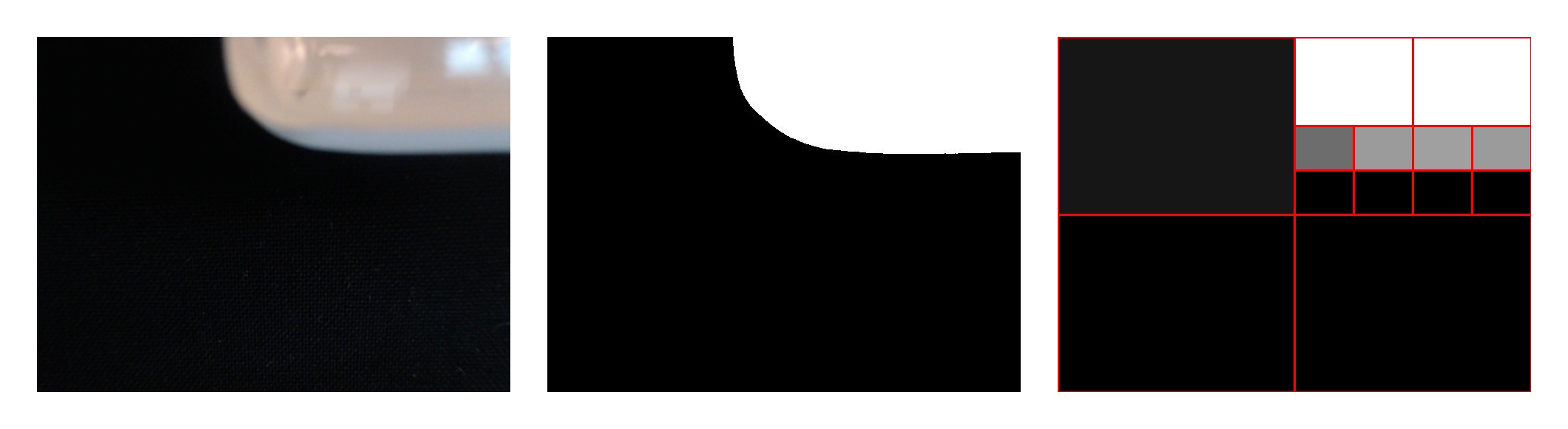}}
		
	\caption{{Examples of inputs and outputs for the obstacle processing unit. (a)-(c): the most left images are the inputs to the unit. The input images are converted to grayscale images first, then a quadtree algorithm is used to optimally partition obstacles within a grayscale images into minimum number of rectangular obstacles  respecting an user-defined threshold for height and width of acceptable smallest rectangle. In these figures, the threshold is 50 for both height and width.}}
	\label{fig:obP} 
\end{figure}

{In the obstacle processor,  a quadtree \mbox{\cite{finkel1974quad}} as a tree data structure is used for spatial 2-D searching to optimally partition obstacles into minimum number of rectangular obstacles within a grayscale image respecting a user-defined threshold for height and width of acceptable smallest rectangle, as depicted in Figure~{\ref{fig:obP}}. As it is shown, there exist regions with different shades of gray and the contrast ranges from black ---pixel value 0--- at the weakest intensity to white ---pixel value 255--- at the strongest. To reduce the cont of computation in the ERRT algorithm, only regions with color intensity of 50 or less.  These rectangular obstacles can  be used efficiently by the ERRT method. }

\subsection{A Discussion on Stability}

{For the stability of this Sub-controller 1 in RL, Attractor dynamics, and} {ERRT-based methods}{, it can be shown that}
\begin{equation}\label{stability_SC1}
{	\begin{split}
		e_1 &= \mathbf{P}_p - \mathbf{P}_{ee},\\
		\dot{e}_1 &= \dot{\mathbf{P}}_p - \dot{\mathbf{P}}_{ee},	\\
		\dot{e}_1 &= \dot{\mathbf{P}}_p - J(q)\dot{q},\\
		\dot{e}_1 &= \dot{\mathbf{P}}_p - J(q)J^{-1}(q)\Big(\dot{\mathbf{P}}_p + K_1(\mathbf{P}_p - \mathbf{P}_{ee})\Big),\\
		\dot{e}_1 &= -K_1e_1
	\end{split}}
\end{equation}
{where $ K_1 $ is a 3-by-3 diagonal positive define matrix, then it is guaranteed that Cartesian vector-valued $ e_1 $ converges to 0 asymptotically for any initial values.

As it is discussed in Section {\ref{RLMethod}}, a closed control architecture within industrial robots or cobots (collaborative robots) such as UR5 has the responsibility of regulating joint angles to desired values (sometimes for simplicity each joint of an industrial arm might be seen as a simple integrator, which is being controlled internally). For the stability of this Sub-controller 2 in RL and} {ERRT-based methods} {, it can be shown that}
\begin{equation}\label{stability_SC2}
{	\begin{split}
		e_2 &= \mathbf{P}_{ref} - \mathbf{P}_{ee},\\
		\dot{e}_2 &= \dot{\mathbf{P}}_{ref} - \dot{\mathbf{P}}_{ee},	\\
		\dot{e}_2 &= \dot{\mathbf{P}}_{ref} - J(q)\dot{q},\\
		\dot{e}_2 &= \dot{\mathbf{P}}_{ref} - J(q)J^{-1}(q)\Big(\dot{\mathbf{P}}_p + K_1(\mathbf{P}_{ref} - \mathbf{P}_{ee})\Big),\\
		\dot{e}_2 &= -K_2 e_2
	\end{split}}
\end{equation}
{where $ K_2 $ is a 3-by-3 diagonal positive define matrix, then it is guaranteed that Cartesian vector-valued $ e_2 $ converges to 0 asymptotically for any initial values.}

{Both RL and ERRT algorithms acts as path planner and it can be seen that robot's internal controller stabilizes the system, as long as an obstacle-free path within workspace of the robot is generated. This is determined by how Sub-controller 2 is well trained in RL algorithm or how good the quality of a path is in ERRT method. Furthermore, in the Attractor dynamic-based method, the global asymptotic stability of Sub-controller 2 is guaranteed as it is discussed earlier in Section {\ref{AttD}}. }

{It should be noted that, each control loop has different set-points and the loops are entirely separated in that sense. In other words the switching between those two stable control loop does not introduce insatiability to the system and the loops accomplish separated sub-tasks with different set-points.  Therefore, this question may raise that whether or not there is a possibility of having only one controller to accomplish the task. Since there is no need to have collision avoidance strategy when the manipulator is moving to the location of disturbed particle, so two different controllers are considered.}

\section{Simulations, Experiments, and Results}

For simulations and real-world experiments, we use the UR5 robotic arm from Universal Robots. The UR5 robot is a collaborative robot with six joints. Built-in safety mechanisms remove the need for safety guards between humans and the robot, and this opens possibilities for using the robot in medical applications where the robot is in direct contact with a patient \cite{mathiassen2016ultrasound}. It has a low-level controller called URControl, and it can be programmed by communicating over a TCP/IP  connection. To control the robot at the script level, a programming language called URScript can be used. URScript programs can be sent from a computer to the robot controller (URControl) as strings over the socket once the connection with the robot is established. The URControl runs programs in real-time and sends information at a {maximum} frequency of 125Hz. The output of the URControl contains all the information the robot needs to perform a movement, including angular positions, velocities, target accelerations, and currents for all joints. In this work, the $movel$ command, which offers a position control interface in tool-space, is used to control the UR5 \cite{mahmood2018setting}. {It should be noted that achieving high-frequency real-time performance for a control loop is limited by the bandwidths of employed sensors and also the computational time required to calculate the control input.}

\begin{figure}[htbp]
    \centering
	\includegraphics[width=0.45\textwidth]{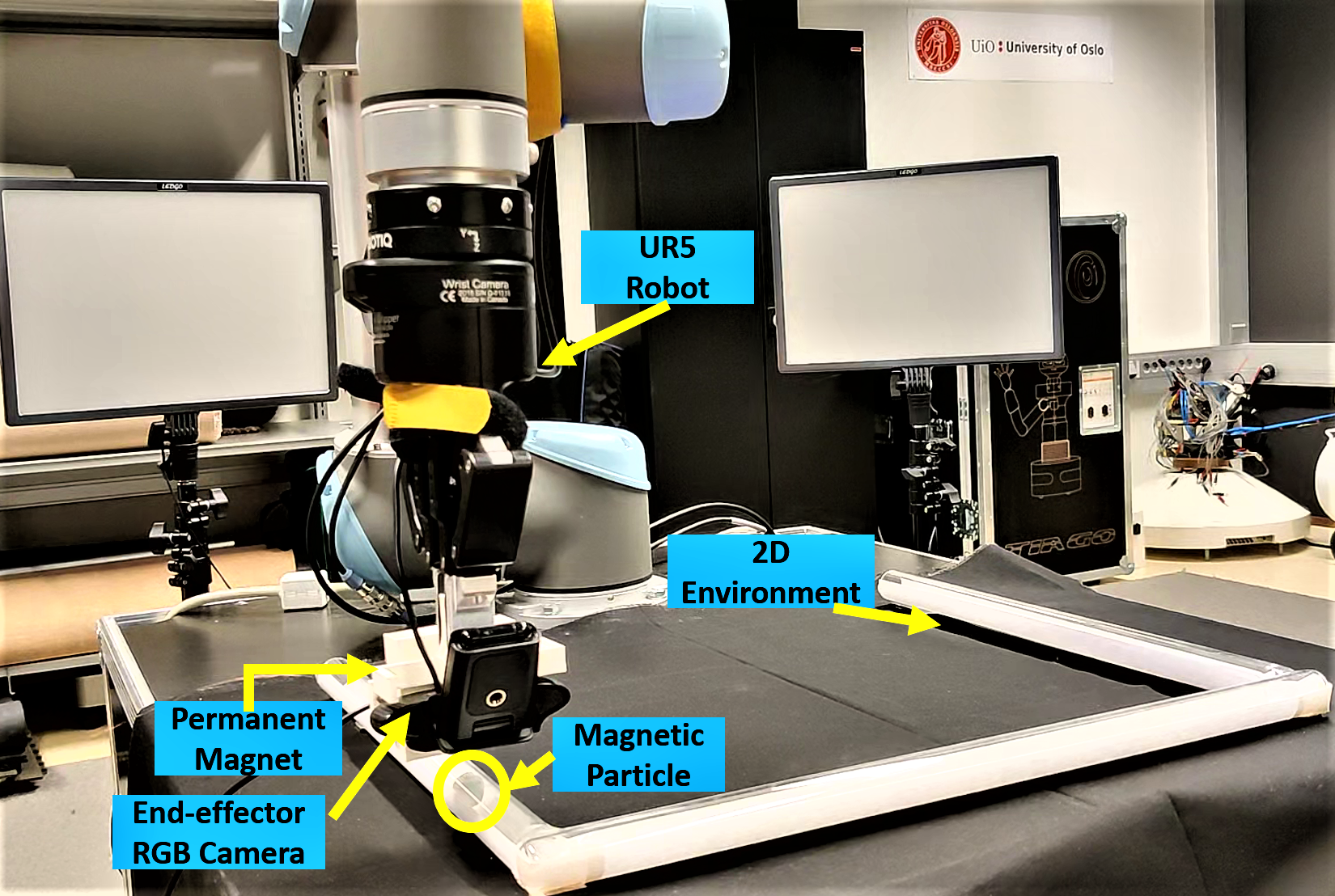} 
    \caption{The experimental setup: a UR5, a permanent magnet, and an RGB camera mounted on the end-effector; the constrained 2D workspace; and the object (a sphere magnet).}
    \label{fig:adaption}
\end{figure}

The UR5 manipulator is equipped with the Robotiq 2F-85 adaptive gripper, holding a neodymium block magnet in a fixed position. The experiment setup is shown in Figure~\ref{fig:adaption}, where images from the end-effector camera are used to detect the magnet and identify obstacles in the environment.

\begin{figure}
	\centering
	%	\subfloat[a\label{1b}]{T
	\subfloat[\label{mag_502510}]{%
		\includegraphics[width=.45\linewidth]{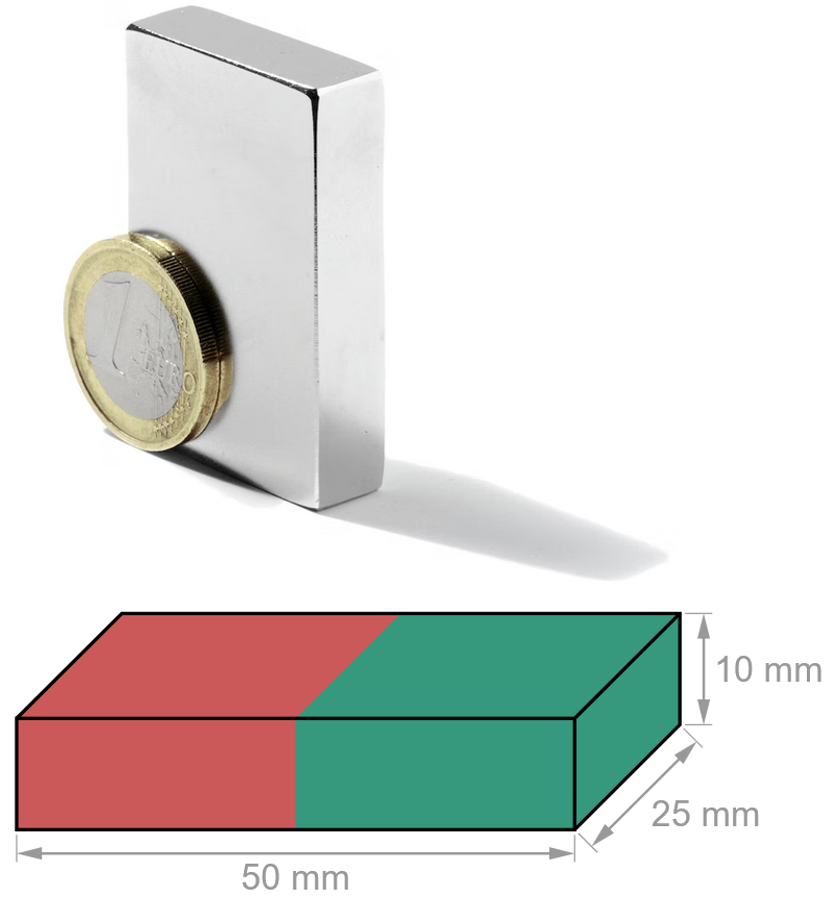}}
	\subfloat[\label{mag_5}]{%
		\includegraphics[width=.325\linewidth]{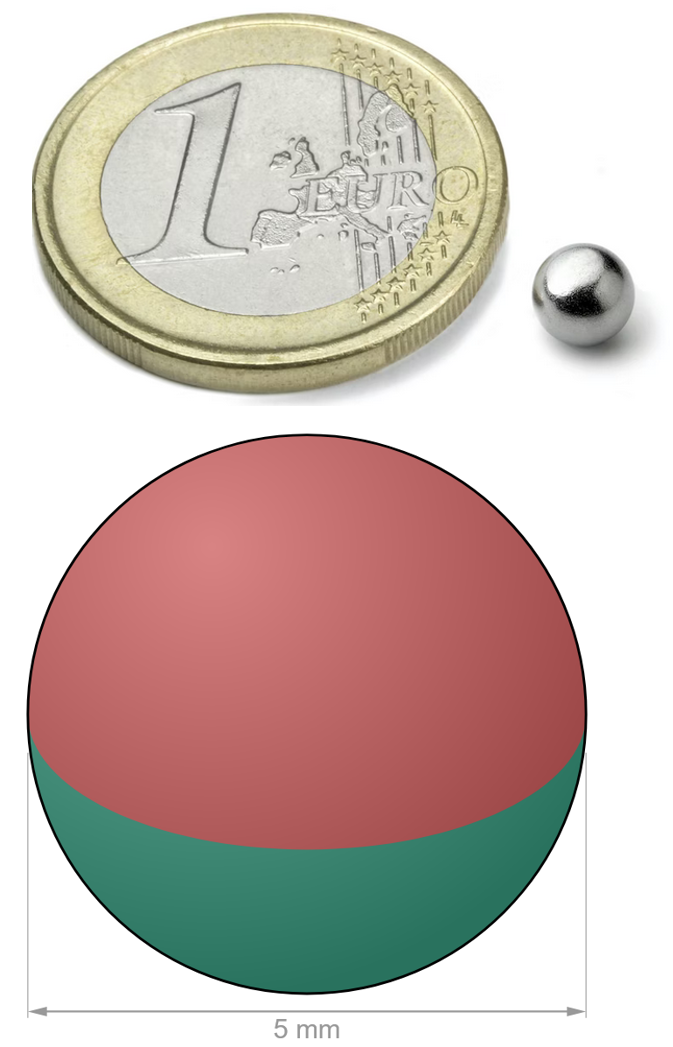}}

	\caption{ (a): A neodymium block magnet with dimensions of $50 \times 25 \times 10$ mm and grade N$35$. (b): A sphere NdFeb magnet with grade N$42$ and diameter $5$ mm as a magnetic particle. }
	\label{fig:magnets} 
\end{figure}

 We use a sphere NdFeb magnet with grade N$42$ and diameter $5$ mm as a magnetic particle. Besides, a neodymium block magnet with dimensions $50 \times 25 \times 10$ mm with grade N$35$ is attached to the UR5's end-effector to produce enough magnetic field strength to carry the magnetic particle. The magnets are shown in Figure~\ref{fig:magnets}. To avoid a collision between two magnets, the UR5  moves in an imaginary plane with a $10$ cm distance from the surface of the table.

\subsection{Simulations, Experiments, and Results: Reinforcement Learning-based Approach}
Training a policy in the real world can be costly. Simulations can speed up the learning process and help avoid potentially unwanted actions that can damage the robot or the surrounding environment. OpenAI Gym and PyBullet \cite{coumans2019} are used for this purpose. However, modeling a complex environment or robots can be challenging, and this may introduce a simulation-reality gap \cite{bousmalis2018using}. To minimize this gap, the UR5 swept and imaged the whole real-world environment using the mounted RGB camera (with step $= 1$ cm in both $x$ and $y$-axes). As discussed, the UR5 end-effector was kept in an imaginary plane with a $10$ cm distance from the surface of the table during the imaging process. Samples of these images are shown in Figure~\ref{fig:train}. The images are fed to the algorithm in the training phase.

\begin{figure}
\centering
    \includegraphics[width=0.45\textwidth]{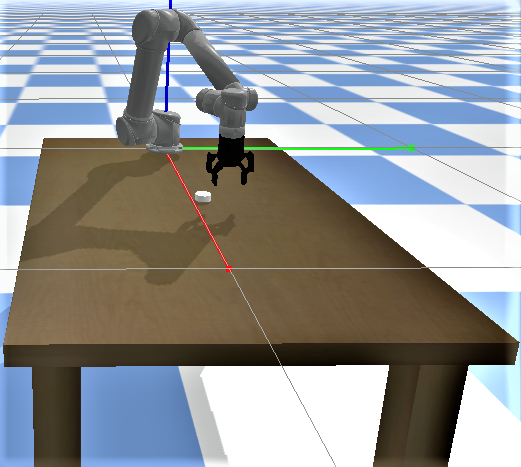}
    \caption{The simulation environment used to train the Rainbow controller: Images taken from the real-world environment are used as observations. The robot gets a negative reward for hitting obstacles and not finding an inverse kinematic solution. It gets positive rewards for reaching the target.}
    \label{fig:env1}
\end{figure}

An overview of the environment used for training Sub-controller 2 is shown in Figure~\ref{fig:env1}. The customized Rainbow algorithm, as previously explained, is employed, where we add images from the end-effector RGB camera together with goal and end-effector positions to the observation space. Image-based observations are used to avoid collisions with obstacles in the environment. Figure~\ref{fig:train} shows details of the training. 

\begin{figure}[h]
    \centering
    \includegraphics[scale = 0.45]{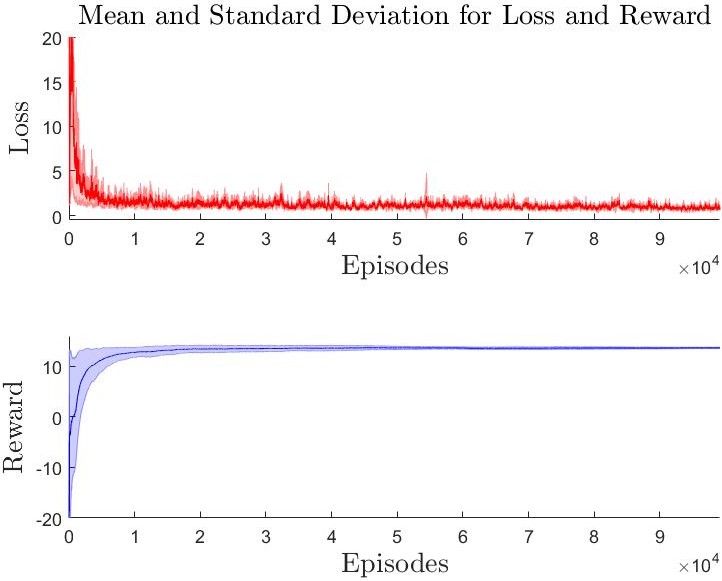}
    \caption{The mean and standard deviation of loss and rewards over each training episode for 5 rounds of training. At the beginning of training, the system's performance is poor, as there is a high standard deviation for the losses and rewards. Eventually, the average and standard deviation of the rewards and losses will improve, meaning that the system will become more reliable and, statistically speaking, will accomplish the task.}
    \label{fig:learningPerf}
\end{figure}

As mentioned, OpenAI Gym and PyBullet are employed to train the customized Rainbow algorithm for Sub-controller 2, and Figure~\ref{fig:learningPerf} represents the learning performance of the employed RL algorithm over the training episodes. It is worth mentioning that in the simulation, the magnetic particle is not considered, and therefore the whole switching control architecture is not implemented. After training, the average success rate for the learned policy is 100\% for randomly generated initial and goal positions (calculated over 30 random restarts).

\begin{figure}
    \centering
    \includegraphics[width=0.5\textwidth]{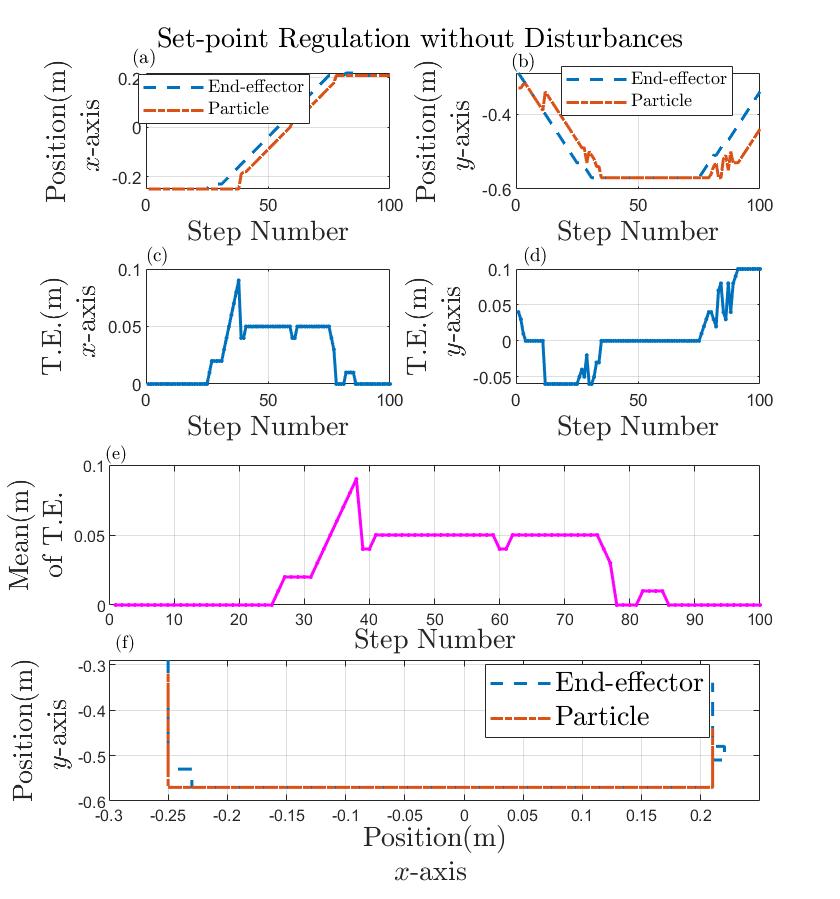}
    \caption{{A scenario of set-point regulation without disturbances in one episode. (a,b): Cartesian positions of the end-effector and the magnetic particle are depicted in both $ x $ and $ y $-axes. (c,d): Tracking Errors (T.E.) in each axis are shown. Errors in each axis increase when there is motion in that axis. (e): The Euclidean norm of errors in both axes at each step number is shown. (f): The locus of the particle and end-effector in the 2-D workspace is depicted.}}
    \label{fig:wod}
\end{figure}

Afterward, we consider the magnetic particle in the loop, and the full controller structure is implemented. In the real world, the magnetic particle is consistently carried to a randomly placed target starting from a random position. We repeated undisturbed scenarios for arbitrary initial and goal positions for 30 episodes, and the average success rate is 100\% {with the mean tracking error 0.02 (m) in both axes, the average maximum error 0.08 (m) and 0.09 (m) in $ x $, and $ y $-axes, respectively.} Figure~\ref{fig:wod} represents a set-point regulation scenario without any disturbances. In this specific scenario, the particle is supposed to reach the goal position $[0.21,-0.40, 0.10]$ within the constrained environment.

\begin{figure}[h]
    \centering
	\includegraphics[width=0.5\textwidth]{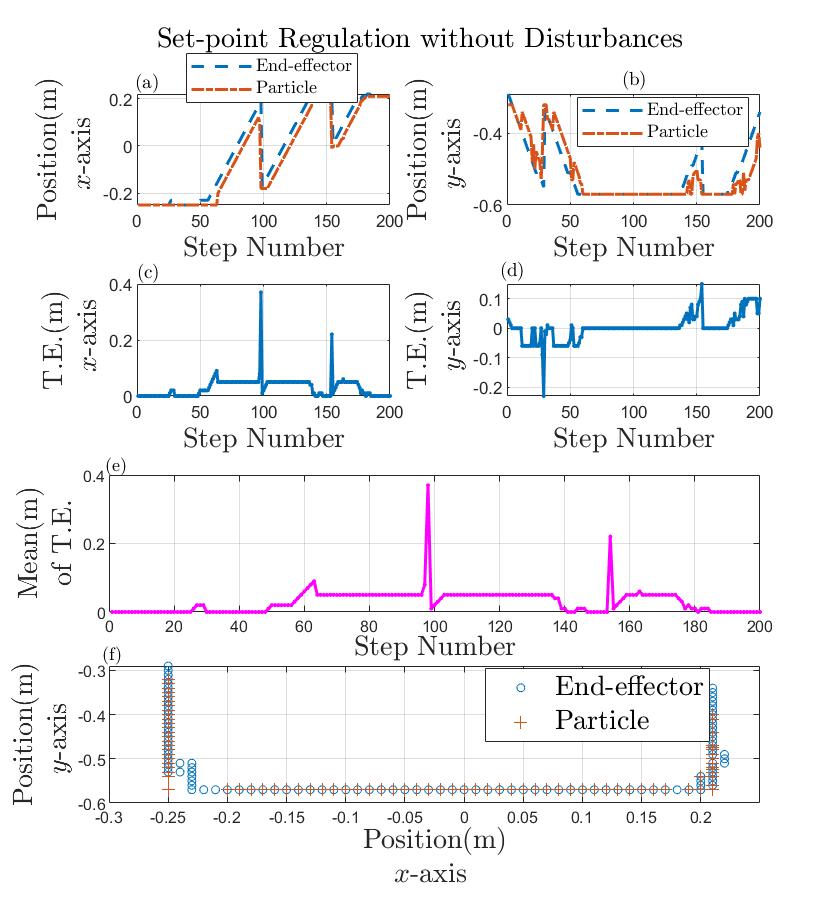}
%    \caption{Set-point regulation with disturbances in one episode.}
    \caption{A scenario of set-point regulation with three disturbances in one episode. Disturbances happen at step numbers 29, 98, and 154 (a,b): Cartesian positions of the end-effector and the magnetic particle are depicted in both $ x $ and $ y $-axes. In (a), disturbances at step numbers 98 and 154, and in (b), a disturbance at step number 98 is visible. (c,d): Tracking Errors (T.E.) in each axis are shown. Errors in each axis increase when there is motion in that axis or when a disturbance happens. The magnitude of disturbances at step numbers 98 and 154 and step number 29 can be seen in (c) and (d), respectively. (e): The Euclidean norm of errors in both axes at each step number is shown. (f): The locus of the particle and end-effector in the 2-D workspace is depicted.}
    \label{fig:wd}
\end{figure}

Finally, the robustness of the system to deal with external disturbances which could result from real-world uncertainties is studied. As disturbances, we physically draw the magnetic particle away from the magnetic control zone during active control. As a demonstration case, Figure~\ref{fig:wd} depicts the particle and end-effector trajectories through the environment in the presence of disturbances where the particle is supposed to reach the goal position $[0.21,-0.40, 0.10]$. We repeated the disturbed scenarios for random disturbances, initial, and target positions for 30 episodes, and the average success rate was 96.6\%.

{In Figure~{\ref{fig:learningPerf}}, we run 5 rounds of training, and each training consists of multiple episodes (more than 90 thousand). In each episode, initial positions and goals are set randomly. In the initial episodes (for all 5 rounds of training), the robot most likely fails to accomplish the task, which is why the average of the rewards for the initial episodes considering 5 rounds of training is a negative number and there is a relatively high loss as well. In addition, it can be seen that at the beginning of training, the system's performance is poor as there is a high standard deviation for the losses and rewards considering the 5 rounds of training. After some more episodes, the average and standard deviation of the rewards and losses, considering all rounds of training, get better, and this means that the system gets more reliable and, statistically speaking, it will accomplish the task. After 90 thousand episodes, the time to reach a random target position from a random initial position will differ. Figures~{\ref{fig:wod}} and {\ref{fig:wd}} only show two episodes after the controller is trained (after 90 thousand episodes).}

\subsection{Simulations and Results: Attractor Dynamics-based Approach}

{The following example illustrates the Attractor Dynamic-based method without any disturbance applied. }
\begin{example}
{Consider a 2-D shape of the constrained environment with the manipulator's initial Cartesian position  $ [-0.78, 0.73, 0.10] $ and target position $ [0.77, 0.19, 0.10] $. An overview of the simulation is shown in Figure~{\ref{fig:simMII}}. It should be noted that the manipulator can sense an obstacle partially with a maximum height and width of 2 cm. The system employs the state feedbaack control law given in {\eqref{ctrlawstate}}. Since there is no perturbation is assumed, Sub-controller 1 (as shown in  Figure~{\ref{fig:attractorDyn}}) is not active.}

\begin{figure} 
	\centering
%	\subfloat[a\label{1b}]{T
	\subfloat[\label{1a}]{%
		\includegraphics[width=0.45\linewidth]{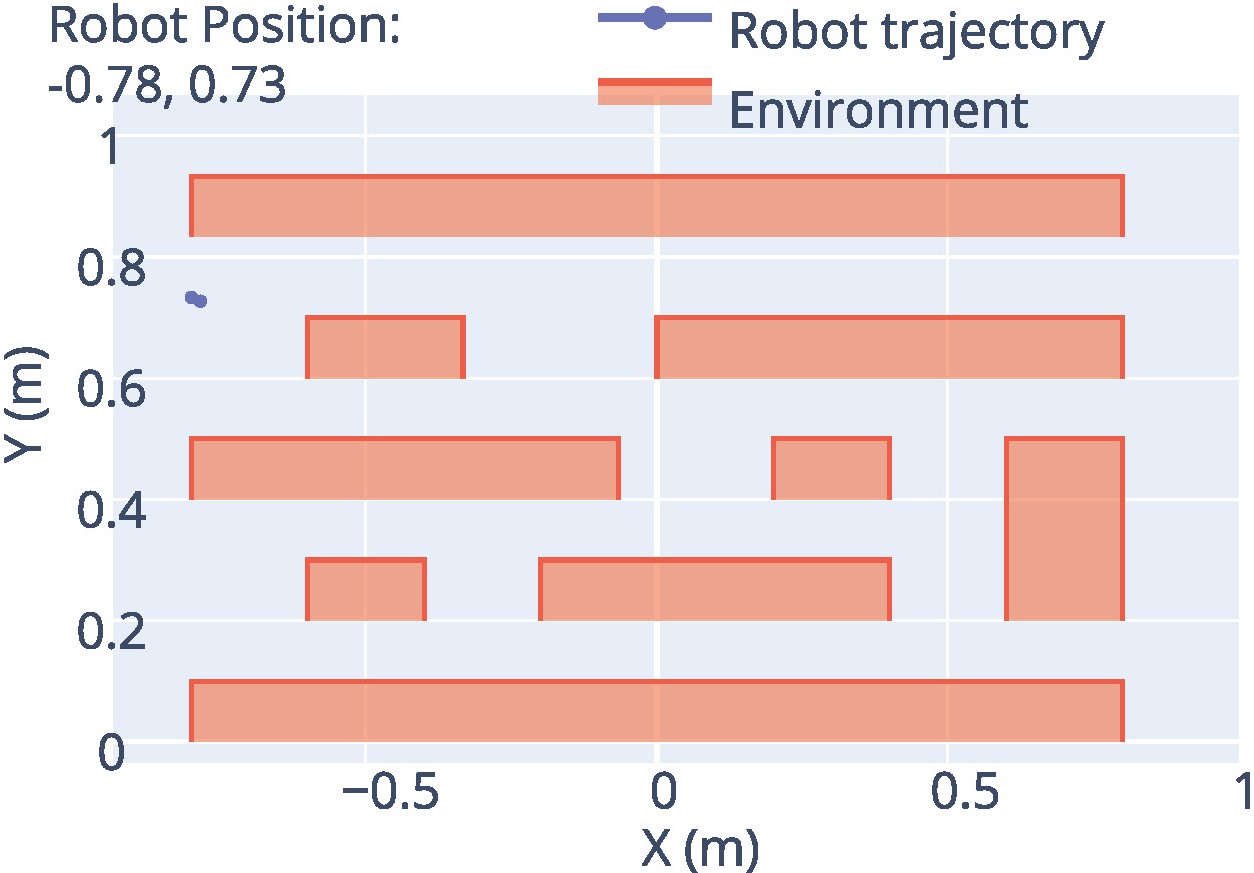}}
	\hfill
	\subfloat[\label{1b}]{%
		\includegraphics[width=0.45\linewidth]{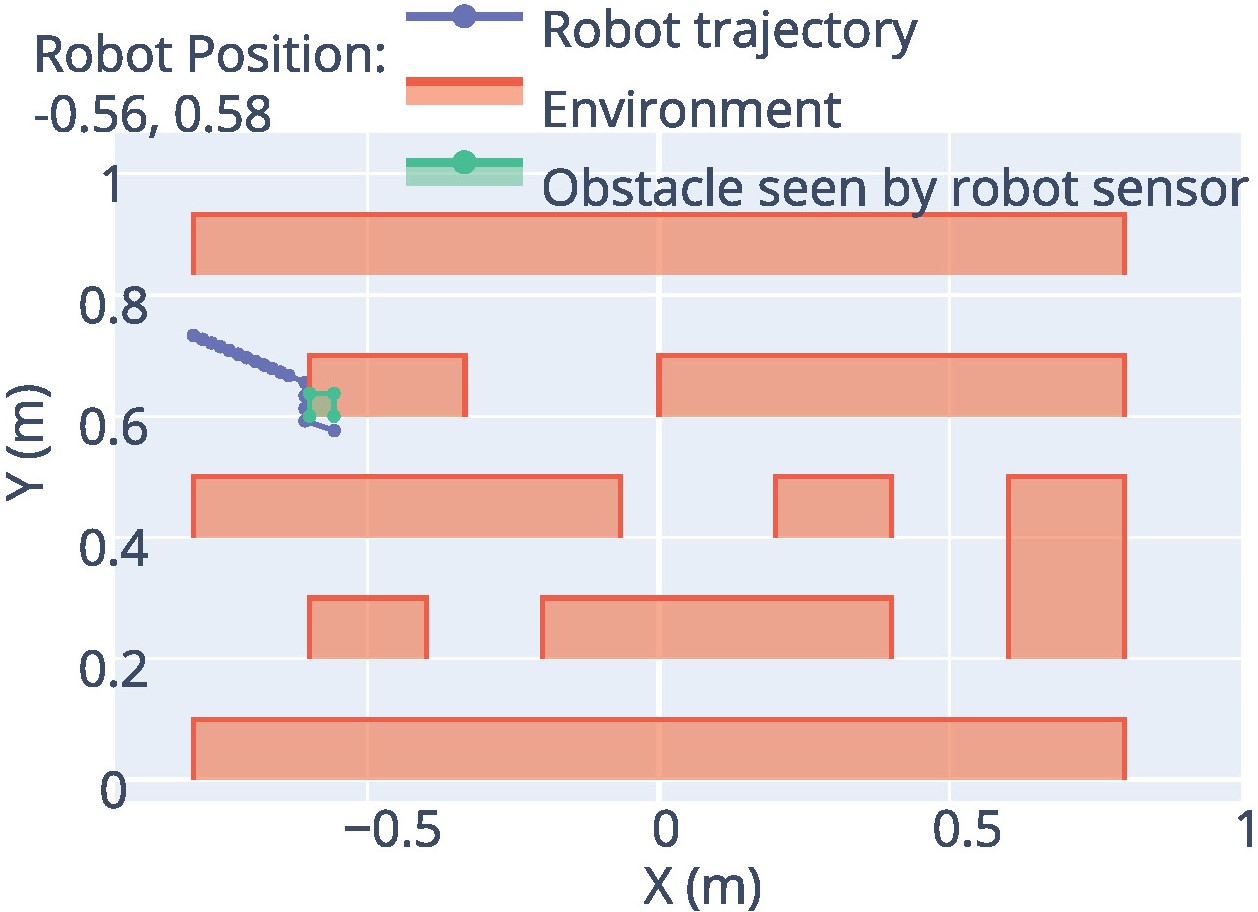}}
	\\
	\subfloat[\label{1c}]{%
		\includegraphics[width=0.45\linewidth]{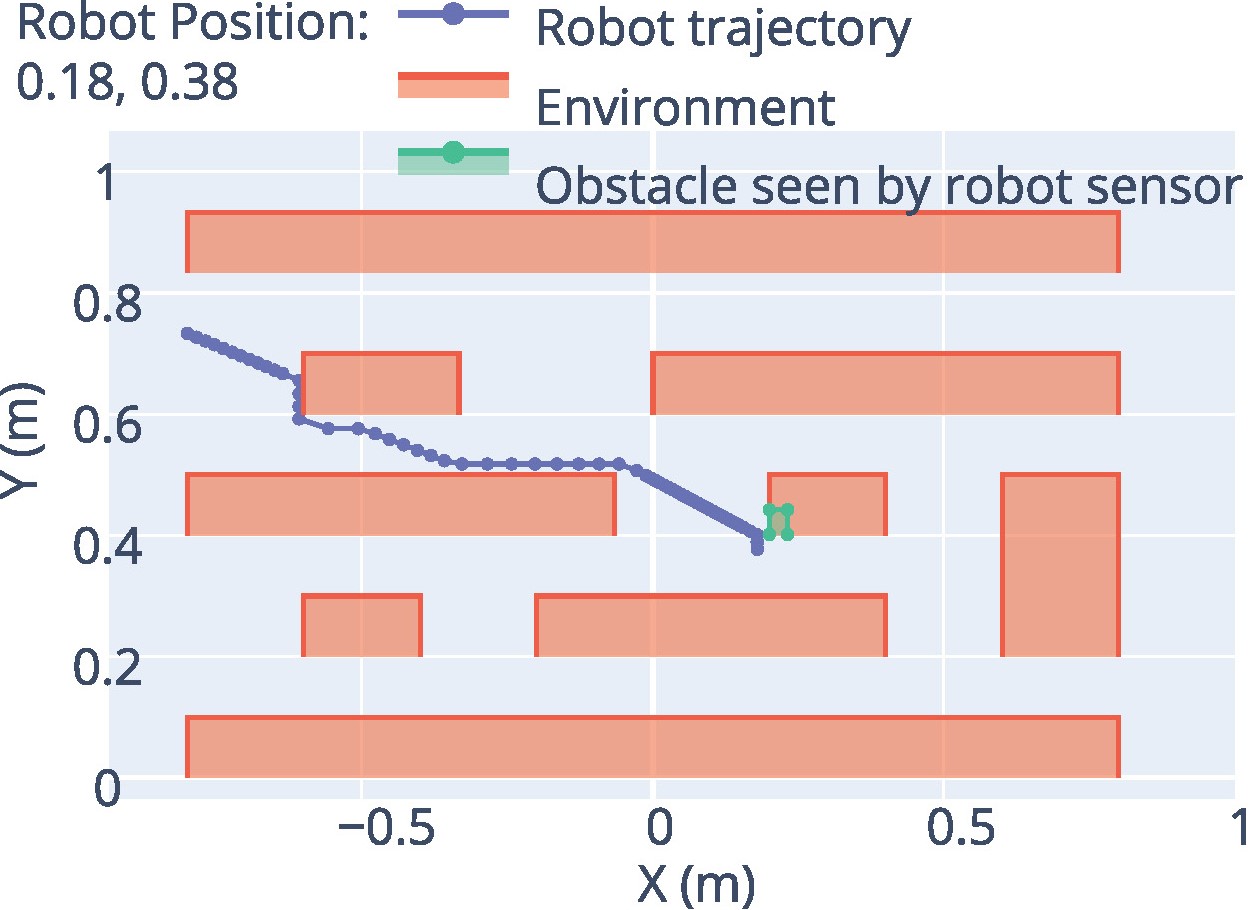}}
	\hfill
	\subfloat[\label{1d}]{%
		\includegraphics[width=0.45\linewidth]{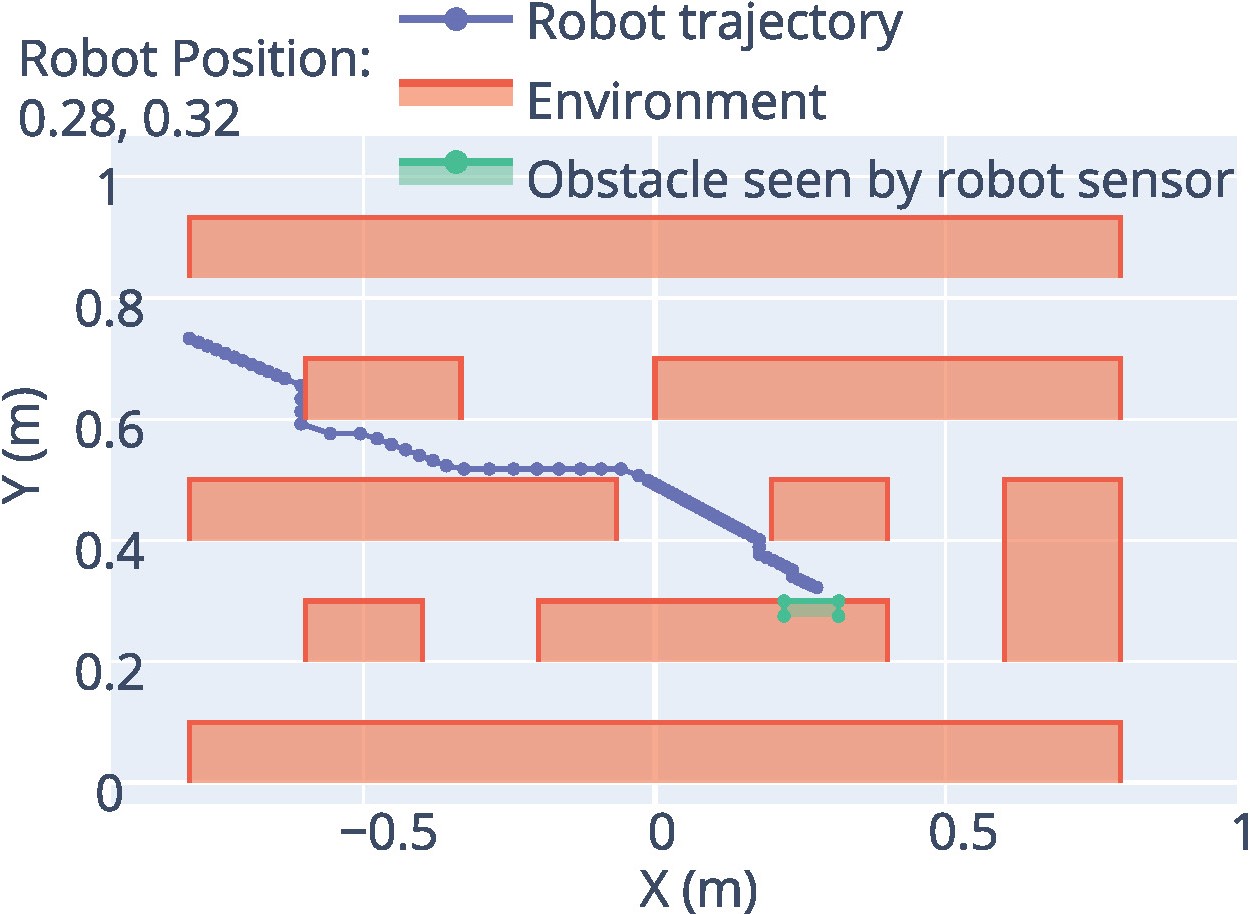}}
	\\
	\subfloat[\label{1e}]{%
		\includegraphics[width=0.45\linewidth]{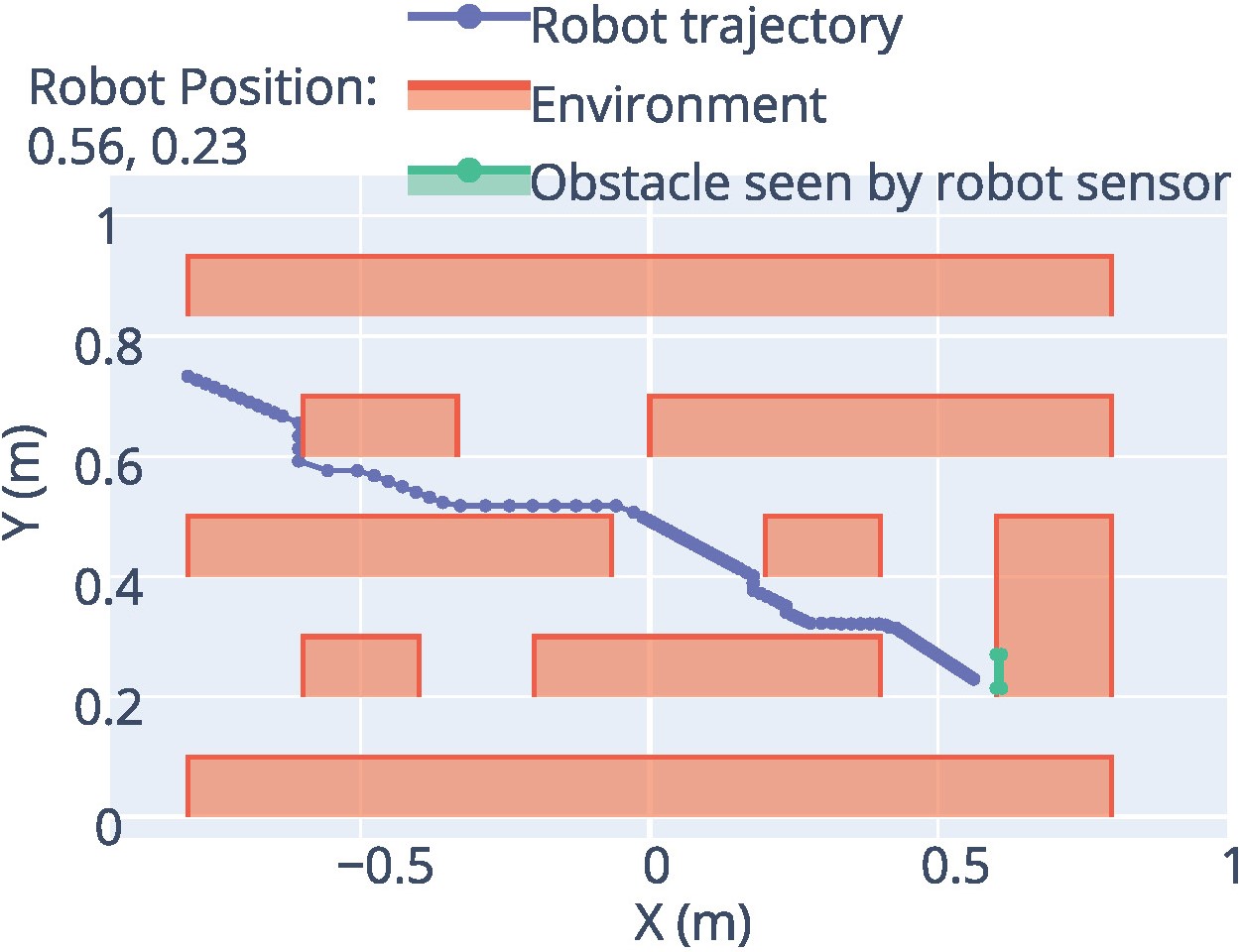}}
	\hfill
	\subfloat[\label{1f}]{%
		\includegraphics[width=0.45\linewidth]{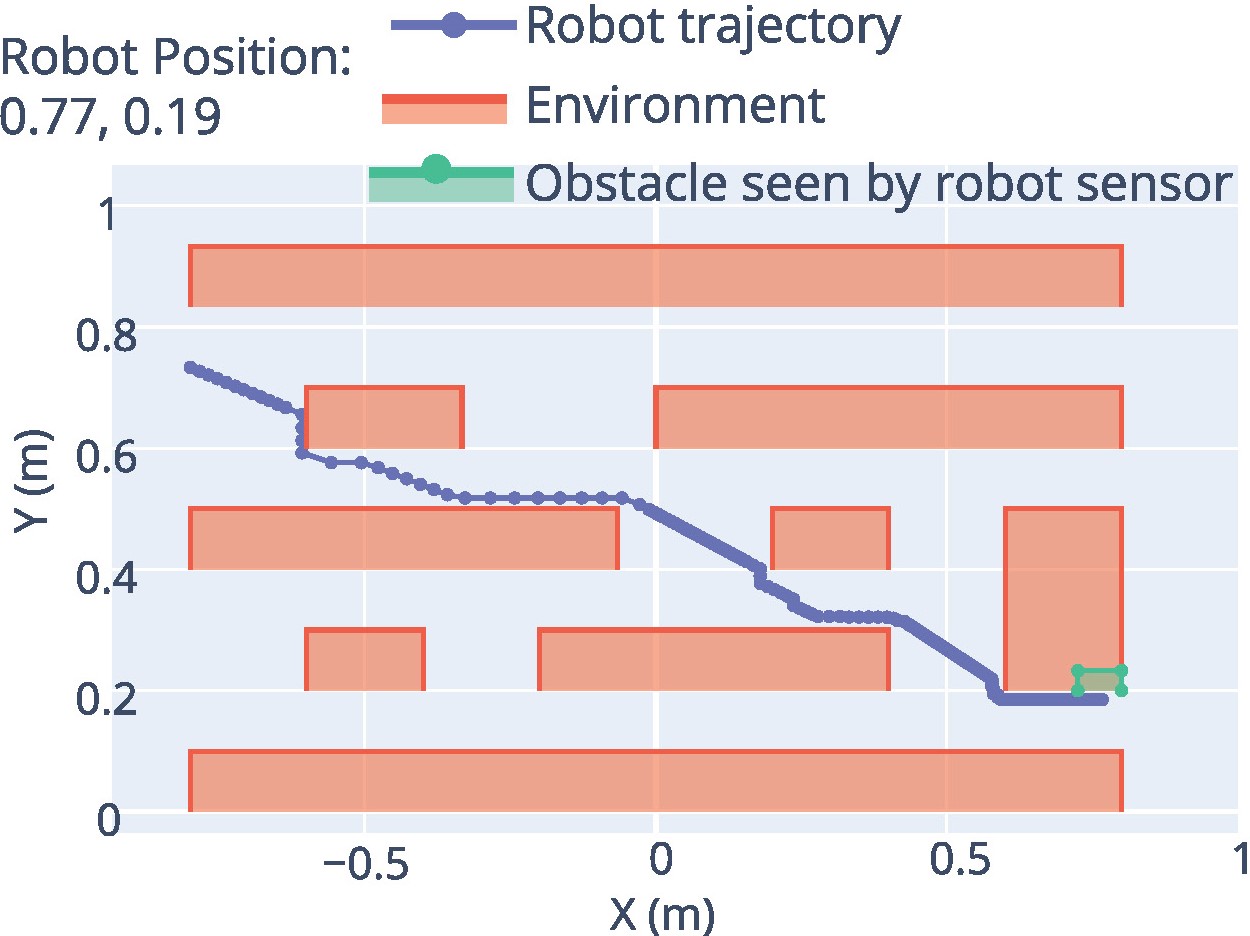}}	
	
	\caption{Online obstacle avoidance of UR5 carrying an undisturbed particle through a 2-D environment as discussed in Example 1. Information from a part of the obstacle seen by the robot is used to make the particle go around the obstacle. Since there is no perturbation, Sub-controller 2 carries the particle toward the target position.} 
	\label{fig:simMII} 
\end{figure}

{An advantage of the considered conventional method is that there is no need for offline training or parameter tuning; however, the method cannot handle concave or connected obstacles. Since the method requires analytical modeling of the obstacles boundary, we consider a rectangle around the part of the obstacle detected by the robot's sensor --- we fit the point
cloud with a rectangle---. However, the actual shape of obstacle(s) may not be
necessarily a rectangle. In other words, the approach might be conservative in some scenarios. }

\end{example}

{In the following examples, a disturbance is introduced to the motion of the particle.}
\begin{example}
{Suppose the same 2-D space as Example 1 is employed. The manipulator's initial Cartesian position  $ [0.72, 0.76, 0.10] $ and target position $ [-0.77, 0.12, 0.10] $. An overview of the simulation is shown in Figure~{\ref{fig:simMII_wd}}. Furthermore, a perturbation in $ x $-axis with magnitude $ 0.63 $ m is applied to the particle when the particle is at the position $ [-0.21, 0.55, 0]$ or at step number 50 where Sub-controller 1 (as shown Figure~{\ref{fig:attractorDyn}}) is active and takes the manipulator to the new position of particle. When the manipulators in the place, Sub-controller 2 has the responsibility of carrying the particle toward the target position. Also in this example, the manipulator can partially sense an obstacle with a maximum height and width of 2 cm.}
\begin{figure} 
	\centering
	%	\subfloat[a\label{1b}]{T
	\subfloat[\label{2a}]{%
		\includegraphics[width=0.45\linewidth, trim={.75cm .8cm 2.0cm 2.1cm},clip]{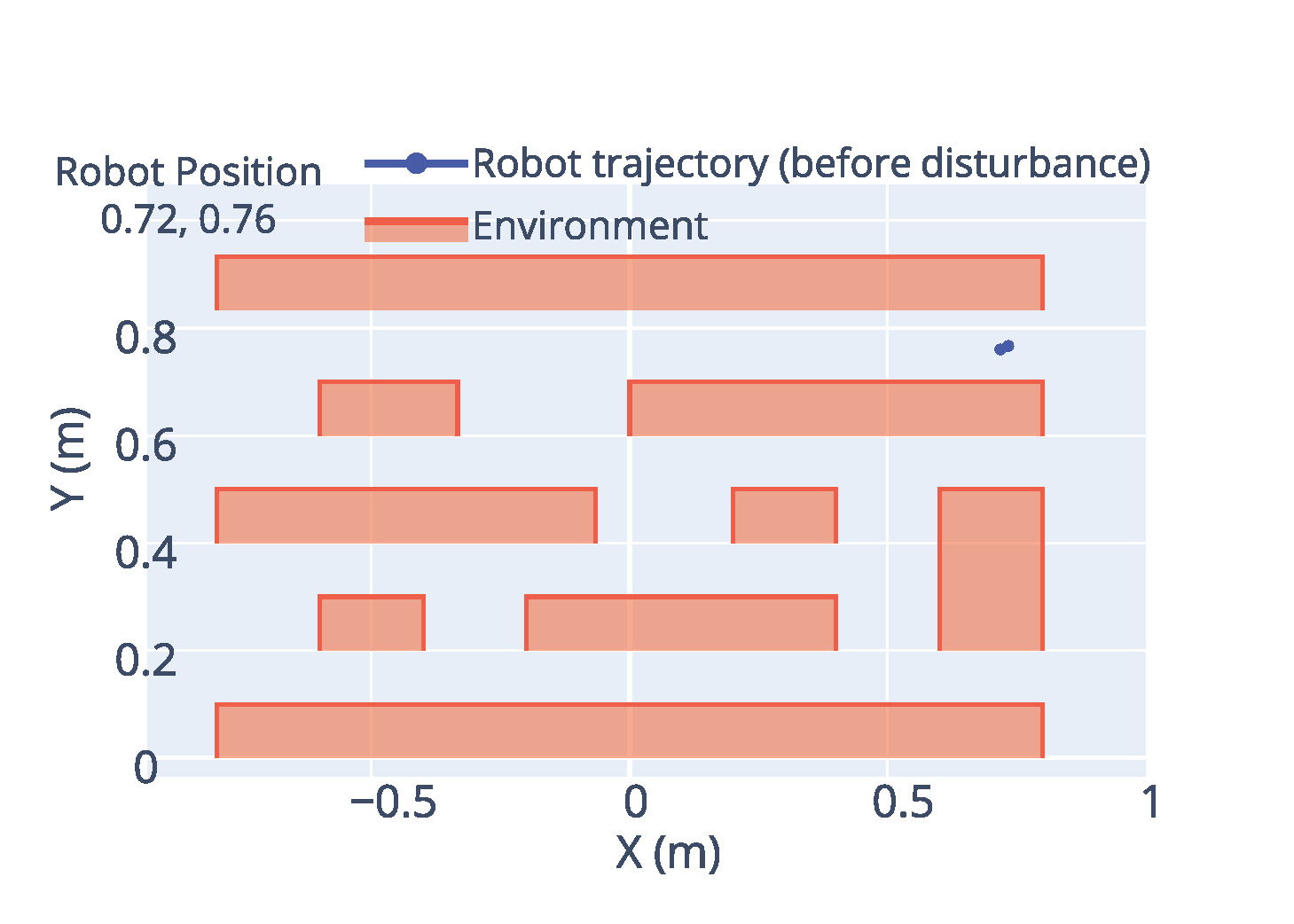}}
	\hfill
	\subfloat[\label{2b}]{%
		\includegraphics[width=0.45\linewidth, trim={.75cm .8cm 2.0cm 1.3cm},clip]{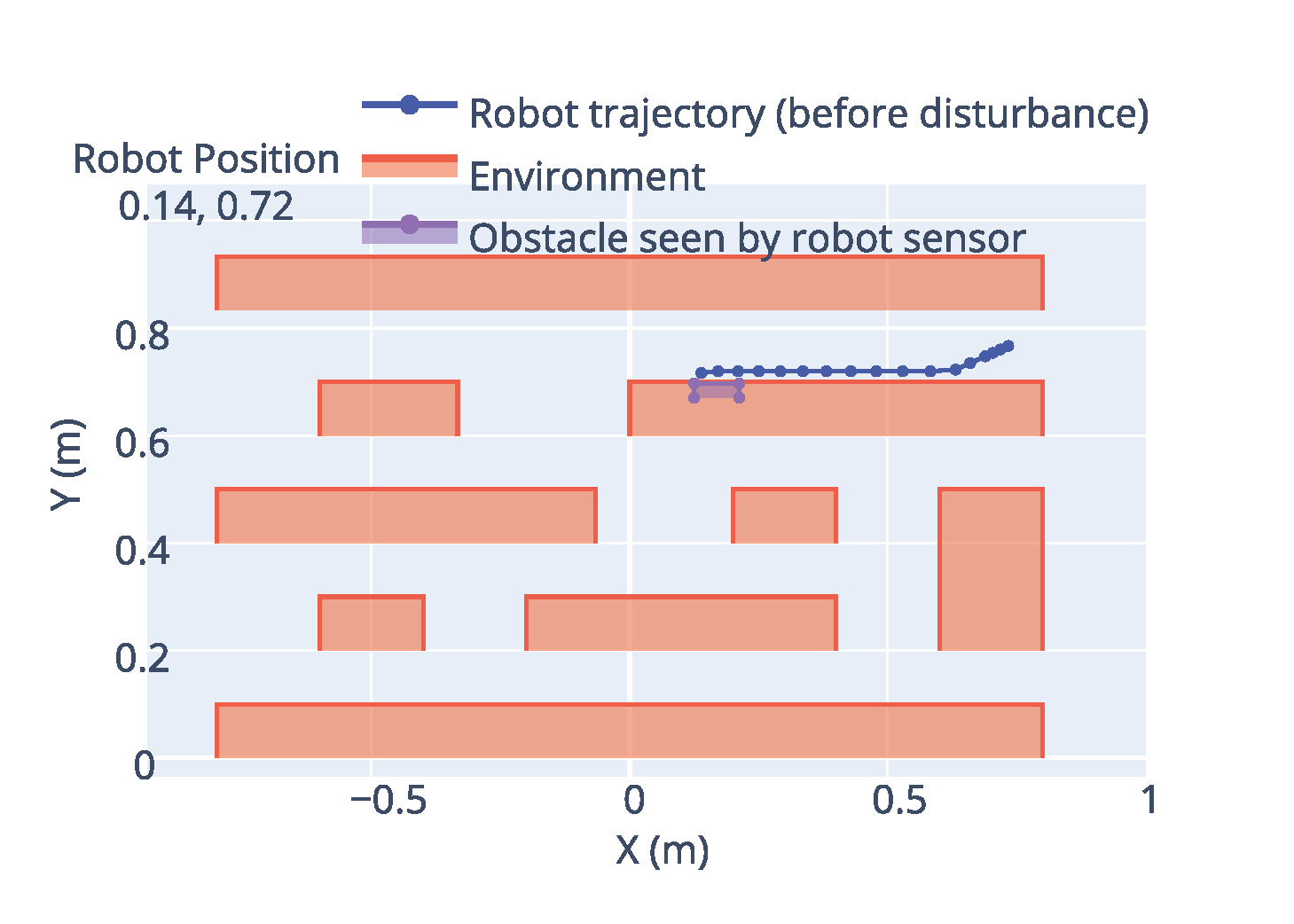}}
	\\
	\subfloat[\label{2c}]{%
		\includegraphics[width=0.45\linewidth, trim={.4cm .8cm 2.0cm 1cm},clip]{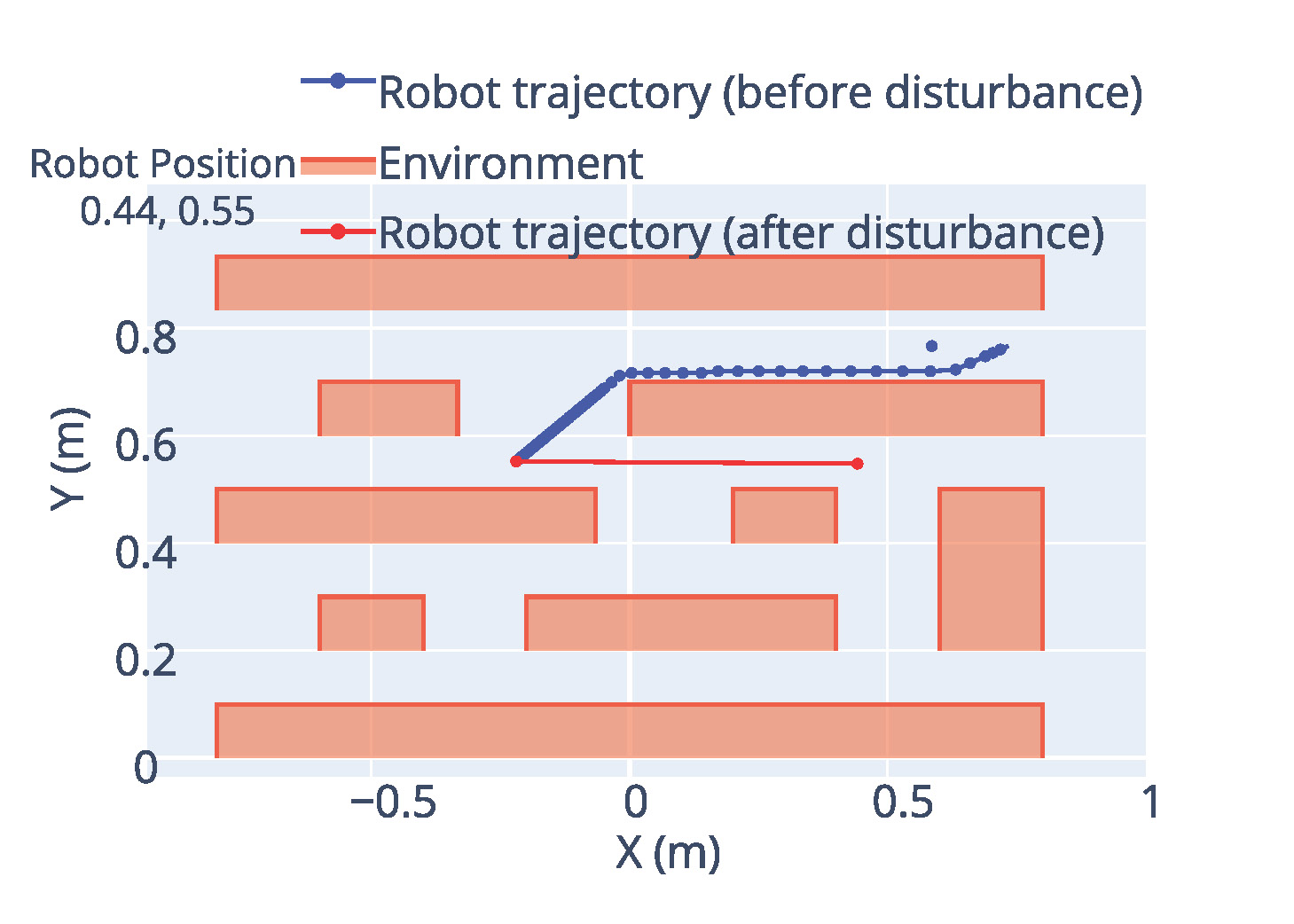}}
	\hfill
	\subfloat[\label{2d}]{%
		\includegraphics[width=0.45\linewidth, trim={.55cm .8cm 2.0cm 0.12cm},clip]{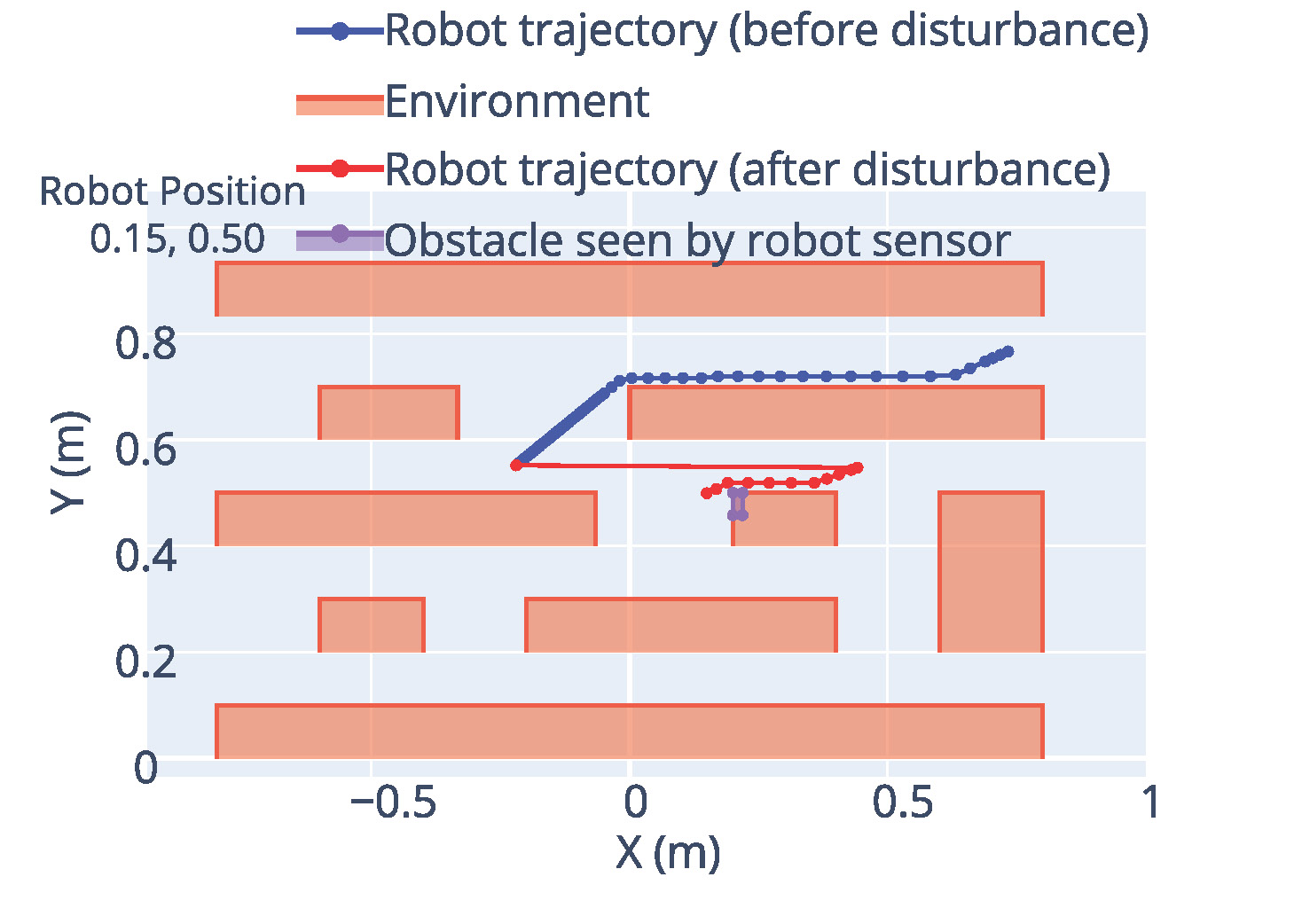}}
	\\
	\subfloat[\label{2e}]{%
		\includegraphics[width=0.45\linewidth, trim={.25cm .8cm 2.0cm 0.08cm},clip]{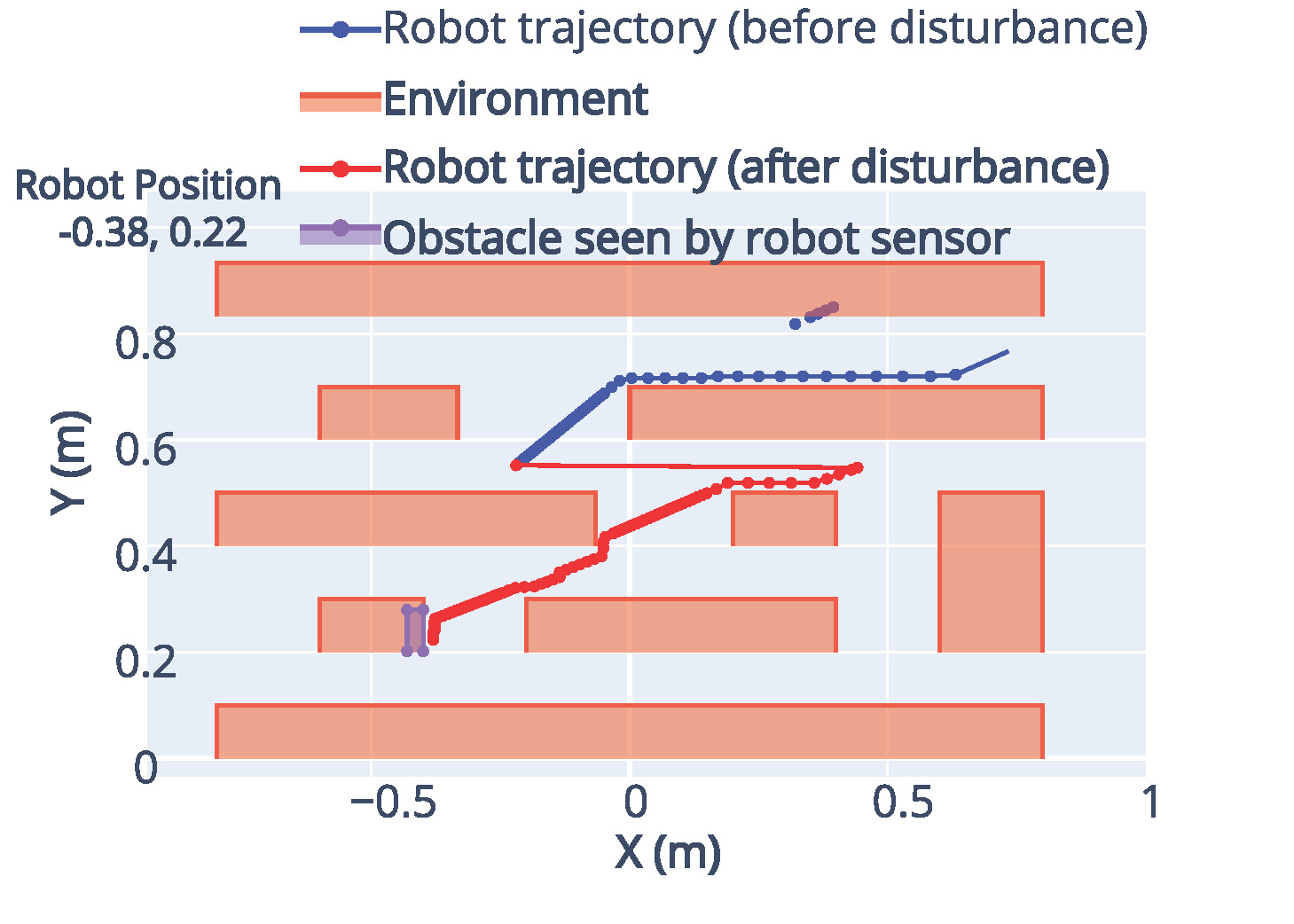}}
	\hfill
	\subfloat[\label{2f}]{%
		\includegraphics[width=0.45\linewidth, trim={.25cm .8cm 2.0cm 0.12cm},clip]{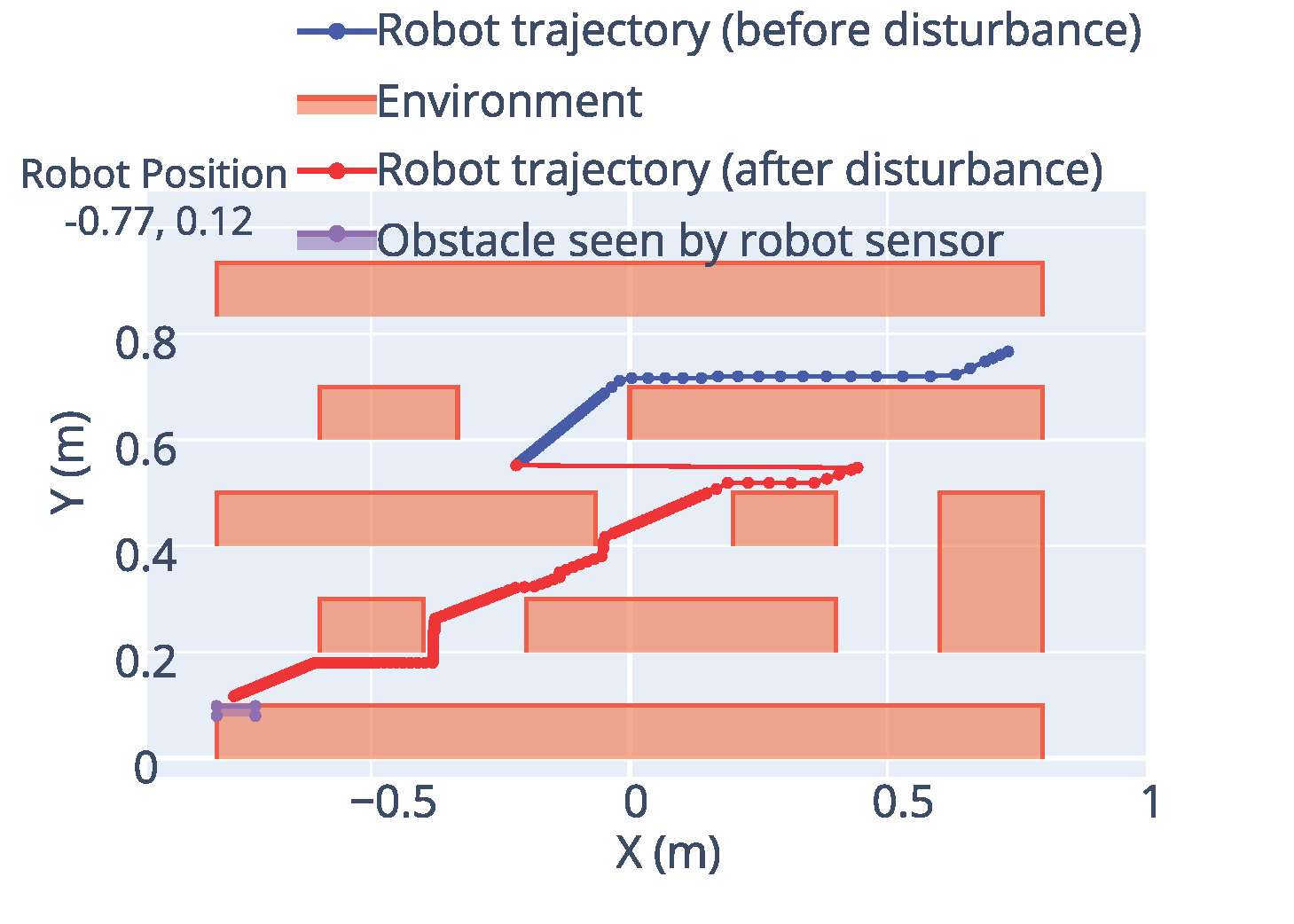}}	
	
	\caption{Online obstacle avoidance of UR5 carrying a disturbed particle through a 2-D environment as discussed in Example 2. A disturbance in $ x $-axis with magnitude $ 0.63 $ m is applied to the particle when the particle is at the position $ [-0.21, 0.55, 0] $, as shown in Figure~{\ref{2d}}. When the disturbance occurs, Sub-controller 1 directs the manipulator to the new position of the particle. When the manipulator reaches the position, Sub-controller 2 has the responsibility of carrying the particle toward the target position (the path shown in red). Information from a part of the obstacle seen by the robot is used to make the particle go around the obstacle.} 
	\label{fig:simMII_wd} 
\end{figure}
{As shown in Figure~{\ref{fig:simMII_wd}}, the particle passes the obstacles
based on received information from the obstacles (part of obstacle seen by the robot), its direction, and relative position to the obstacles.}
\end{example}

\subsection{Simulations and Results: ERRT-based Approach}
{The following example illustrates the performance of  ERRT-based method with disturbances applied to the particle. }
\begin{example}
	
{Consider a 2-D environment as depicted in Figure~{\ref{fig:errt_initial}} with the manipulator's initial Cartesian position  $ [-0.24, -0.27, 0.10] $ and target position $ [0.19, -0.28, 0.10] $.}

\begin{figure}
	\centering
	%	\subfloat[a\label{1b}]{T
	\subfloat[\label{errt_Figure_1}]{%
		\includegraphics[width=0.85\linewidth, trim={0.15cm 0cm 1.55cm 0.9cm},clip]{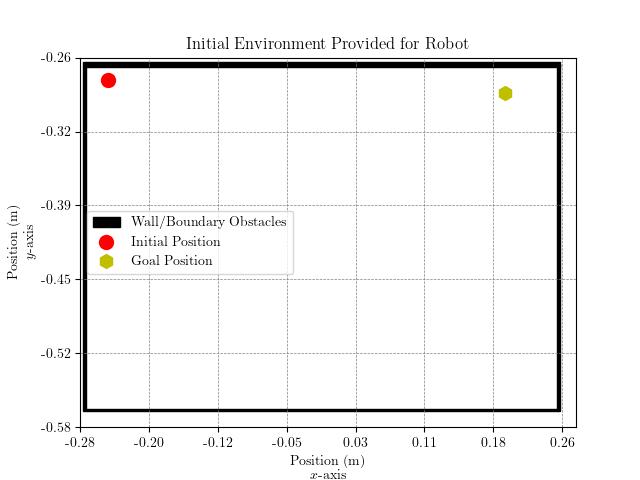}}
	\\
	\subfloat[\label{errt_Figure_2}]{%
		\includegraphics[width=0.85\linewidth, trim={0.15cm 0cm 1.55cm 0.5cm},clip]{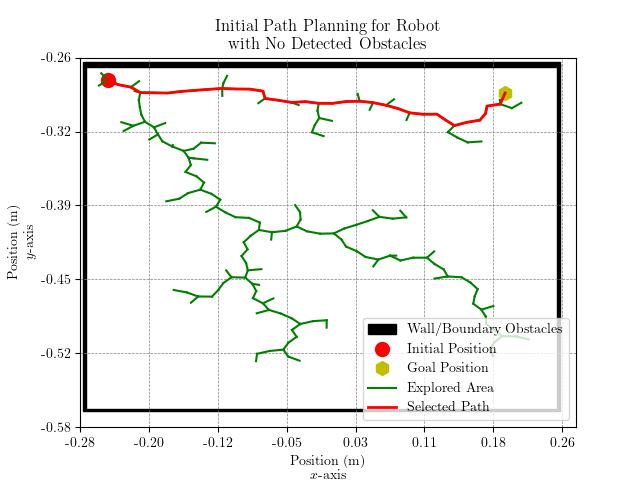}}
	
	\caption{(a): The initial environment provided for the robot with boundary obstacles to limit the search space. (b): The initial planned path by ERRT-based path planner from the initial position to the goal position together with the explored and selected paths and boundary obstacles to limit the search space. No obstacle is detected by the camera.}
	\label{fig:errt_initial} 
\end{figure}

{In the next scenario, two disturbances happen at locations $ [0.21, -0.34, 0.10] $ and $ [0.20, -0.47, 0.10] $ where the particle is displace to new positions $ [-0.19, -0.48, 0.10] $ and $ [-0.24, -0.28, 0.10] $, respectively as shown in Figure~{\ref{fig:traject_errt}} where initial, target, and disturbance positions, explored area and selected path are depicted. In addition, it should be noted that the ERRT algorithm senses obstacles through received images from the environment depending on the robot's end-effector current position (obstacles are partially seen by the camera mounted on robot).}

\begin{figure}
	\centering

	\subfloat[\label{errtDist_b}]{%
		\includegraphics[width=.85\linewidth, trim={0cm 0cm 1.55cm 0cm},clip]{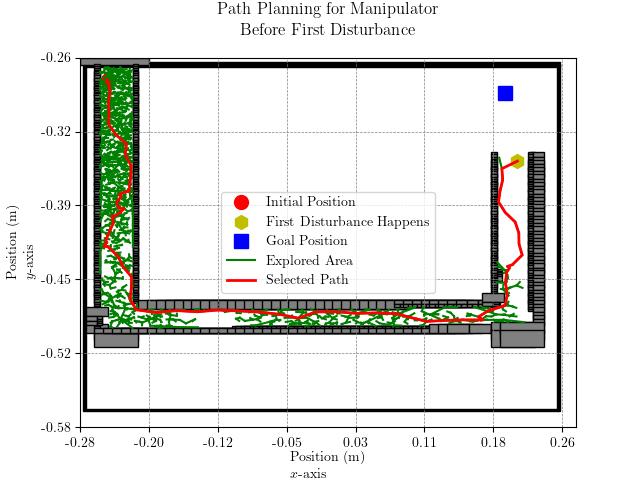}}
	\\
\subfloat[\label{errtDist_c}]{
	\includegraphics[width=.85\linewidth, trim={0cm 0cm 1.55cm 0cm},clip]{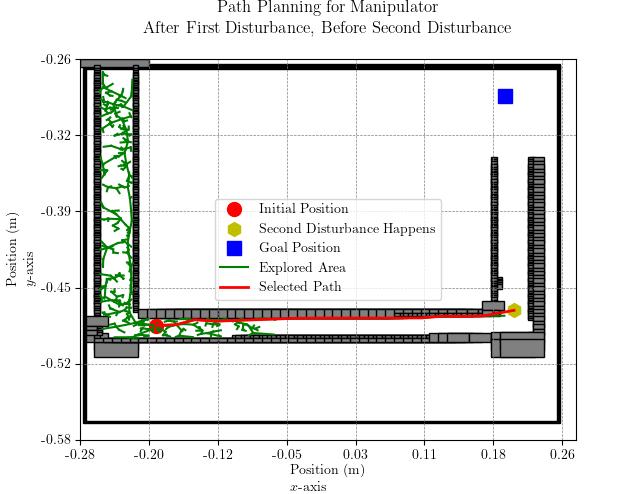}}
\\
\subfloat[\label{errtDist_d}]{%
	\includegraphics[width=.85\linewidth, trim={0cm 0cm 1.55cm 0cm},clip]{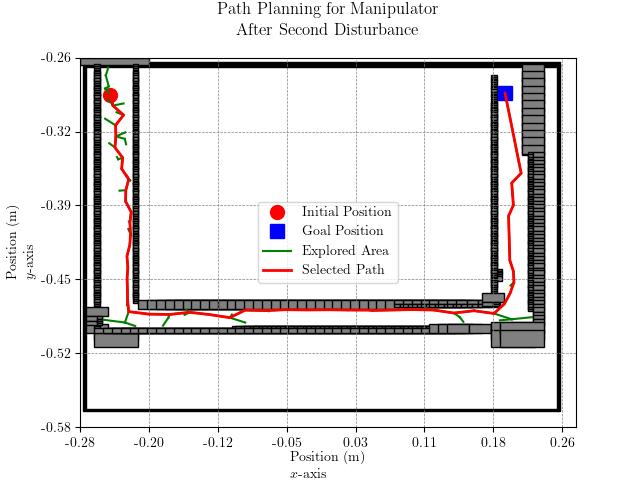}}

	\caption{{The initial, target, and disturbance locations, explored area and selected path. (a): The initial position before the first disturbance is at $ [-0.24, -0.27, 0.10] $, target position is at $ [0.19, -0.28, 0.1] $, and the first disturbance occurs at $ [0.21, -0.34, 0.1] $. (b): The initial position after the first disturbance is at $ [-0.19, -0.48, 0.1] $, the target position is the same as before, and the second disturbance occurs at $ [0.20, -0.47, 0.1] $. (c): The initial position after the second disturbance is at $ [-0.24, -0.28, 0.1] $, the target position is kept the same as before, and no disturbance occurs.}}
	\label{fig:traject_errt} 
\end{figure}

{Figure~{\ref{fig:wd_eert}} depicts the particle and end-effector trajectories together with Tracking Error (T.E.) in both axes $ x $ and $ y $ and mean of T.E., through the environment in the presence of disturbances where the particle is supposed to reach the goal position $[0.21,-0.40, 0.10]$. }

\begin{figure}[h]
	\centering
	\includegraphics[width=\linewidth, trim={.4cm .6cm 2.25cm .75cm},clip] {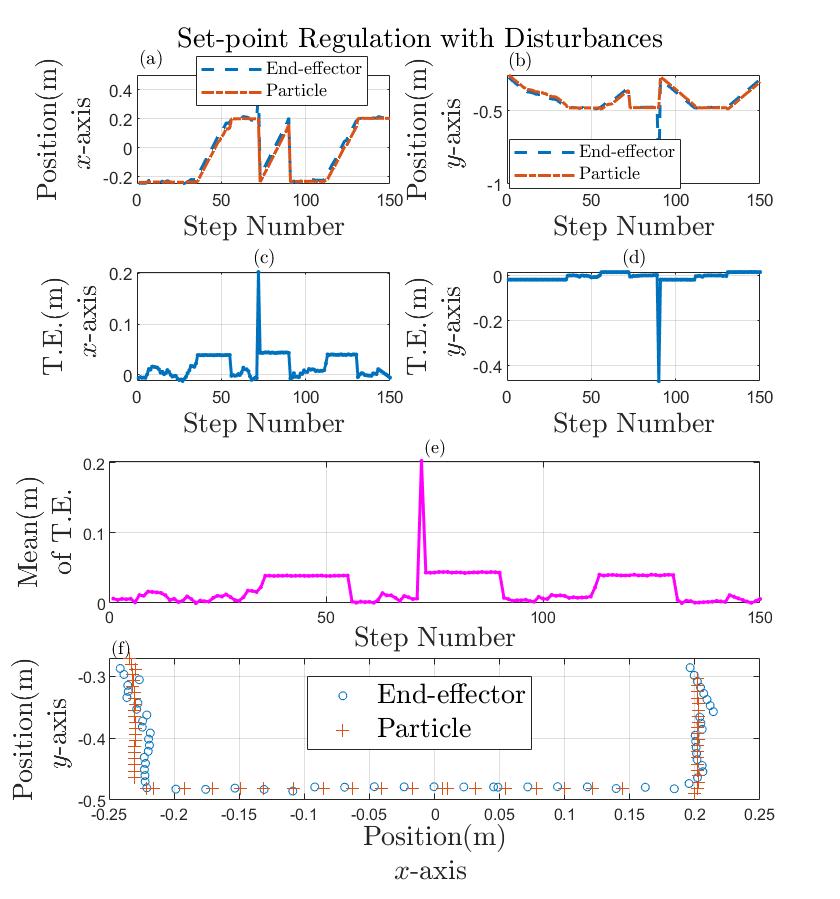}
	%    \caption{Set-point regulation with disturbances in one episode.}
	\caption{{A scenario of set-point regulation with three disturbances in one episode. Disturbances happen at step numbers 72 and 90 (a,b): Cartesian positions of the end-effector and the magnetic particle are depicted in both $ x $ and $ y $-axes. In (a), a disturbance at step numbers 72, and in (b), a disturbance at step number 90 are visible. (c,d): Tracking Errors (T.E.) in each axis are shown. Errors in each axis increase when there is motion in that axis or when a disturbance happens. The magnitude of disturbances at step numbers 72  and step number 90 can be seen in (c) and (d), respectively. (e): The Euclidean norm of errors in both axes at each step number is shown. (f): The locus of the particle and end-effector in the 2-D workspace is depicted.}}
	\label{fig:wd_eert}
\end{figure}

\end{example}
\section{Discussions}

In this paper, we examine the feasibility of implementing the Rainbow algorithm \cite{hessel2017rainbow} customized with Quantile Huber loss from Implicit Quantile Networks (IQN) algorithm\cite{dabney2018implicit} and ResNet\cite{he2016deep} to accomplish set-point regulation tasks within a constrained environment in the presence of disturbances. 
Our results suggest that deep RL methods as a trajectory planner prevent the robot from encountering singularities and guide the particle toward a goal while avoiding obstacles. Furthermore, the results show that it is possible to train the RL algorithm fully within a simulation environment and deploy it as-is in a real-world scenario for remote magnetic control with reliable behavior in the presence of disturbances and uncertainties. Although traditional control approaches may ensure some performance levels under restricted assumptions, their design is challenging, especially when only partial information from the environment is available at each time step.
%In other words, there would be a need for exploration strategies in the design of traditional path planners. 
Also, in many traditional path planning and control design approaches, an exact analytical model of the robot and its interaction with the surrounding environment is a necessity; contrastingly, RL methods do not rely on pre-existing knowledge or models.

{ The drug delivery example is a futuristic, potential application in which the human intervascular system is a 3-D, time-varying, and dynamic environment. However, compared to traditional methods, the strength of RL-based methods is that they are applicable to the more complex environment without a need to re-design the controller structure ---which is based on neural networks---. Yet, there might be a need for more samples in the offline training and proper tweaking of the controller's hyperparameters. }

A simple image segmentation algorithm was employed in this work; however, equipping the RL method with a high-fidelity segmentation algorithm would help in better understanding the environment, thereby reducing the simulation-reality gap and augmenting the generalization of the learning algorithm to be used in a different environment. We experienced that learning performance is also sensitive to the choice and format of observation space; therefore, learning parameters must be chosen carefully.

It should be noted that as a first step to evaluate the feasibility of the proposed approach, we considered an optical tracking system to detect and guarantee the presence of a magnet in the immediate vicinity of the UR5 end-effector and an RGB camera to detect the constrained workspace. However, for more practical uses, all the imaging modalities are better integrated at the end-effector and replaced by more relevant radiography methods, which are considered safe procedures, and can be used to see through the body.

It is worth mentioning that the presented Attractor Dynamics-based conventional method can only be used where obstacle(s) are convex and are not connected. The method requires analytical modeling of the obstacle's boundary. To this end, we consider a rectangle around the part of the obstacle detected by the robot's sensor ---we fit the point cloud with a rectangle---. However, the actual shape of the obstacle(s) may not necessarily be a rectangle. In other words, the approach might be conservative in some scenarios.   
Also, integration errors can cause issues in the discrete implementation of the approach in a way that the next computed point falls inside an obstacle boundary due to the integration error, and trajectories tend to stay inside the obstacle (as no trajectory can enter or leave obstacles).

{To highlight the benefits of the proposed RL-based algorithm, an ERRT-based approach is also employed to accomplish the same task. ERRT is a sampling-based planning method that explores the environment and plans an obstacle-free path on the fly. 

In order to handle feeding the algorithm with varying obstacle shapes, a technique is developed to optimally partition obstacles into rectangular shapes that can be processed by the ERRT algorithm. This is the first difference between RL and ERRT-based methods. In the RL-based approach, there is no need to be concerned about the shape of obstacles as the whole obstacle image will be taken into account. Therefore, by approximating obstacles, the available workspace for generating an obstacle-free path might be reduced. At the same time, this approximation will also increase the computation loads to some extent, which eventually will lead to decreased bandwidth for the control loop.} 

{In the proposed RL-based algorithm, training or exploring the environment takes place offline, yet in the ERRT method, environment exploration happens on the fly, and this directly affects whether the method can be used in real-time. If the environment is filled with complex and/or dynamic obstacles, the ERRT algorithm might not be responsive to the changes due to computational loads. However, this is not the case in the RL-based algorithm. In the presented scenario in this paper, the environment is static so it was feasible to employ ERRT method.

In the ERRT method, there is a need for calibration of the environment with respect to the robot to keep track of the obstacles' positions. Yet, in the RL-based approach, the current pose of the robot will be assigned to the received obstacle image, and this will be preserved in the neural network, which will enable employing the RL-based method in dynamic 3-D environments without being concerned about the calibration. The calibration constraint prevents ERRT from being implemented in 3-D environments. }  

\section{Conclusion}
This article derives and validates a customized Rainbow RL method for online trajectory planning and remote control of a ferromagnetic particle. Using magnetic actuation, the robot learned to robustly carry a small ferromagnetic object in a constrained environment. Furthermore, the trained network is integrated with the two-module controller, which is deployed as-is in a validation experiment in the real world, where the experiment showed the robustness of the approach against disturbances. Also, a conventional controller based on the Attractor Dynamics-based approach is designed. Afterward, simulations were carried out using more complex environments, and the shortcomings of the proposed method were also discussed. Finally, the experiment results from the ERRT-based method highlight the improved robustness to dynamic environments offered by the proposed RL-based algorithm in this work. For future work, a three-dimensional environment can be investigated. Also, in this work, the constrained environment is static, meaning it does not change over time. However, by training the RL algorithm in dynamic environments, there is a possibility of considering time-varying workspaces.

\section*{Acknowledgment}

This work was partially supported by the Research Council of Norway (RCN) as a part of the Vulnerability in the Robot Society (VIROS) project under Grant Agreement No. 288285, Predictive and Intuitive Robot Companion (PIRC) under Grant Agreement No. 312333; and through its Centres of Excellence scheme, Project No. 262762. 

\bibliographystyle{ieeetr}
\bibliography{refs}

% \clearpage

 \begin{IEEEbiography}[{\includegraphics[width=1in,height=1.1in,clip,keepaspectratio]{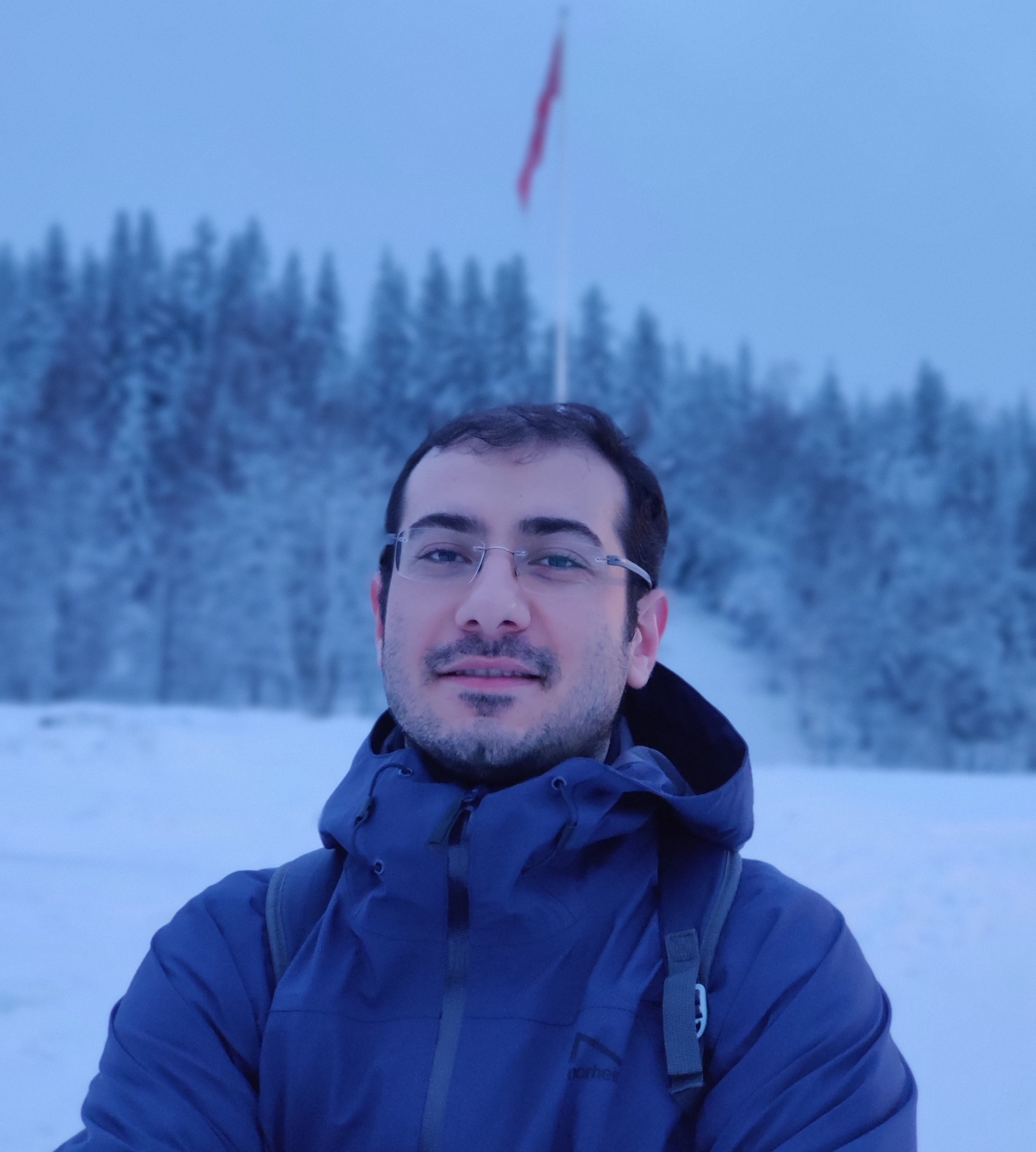}}]{Abbas Tariverdi}
 received the B.Sc. and M.Sc. degree in
 electrical and control engineering from Amirkabir University of Technology (Tehran Polytechnic), Tehran,
 Iran, in 2014 and 2016, respectively. He is currently pursuing the Ph.D. degree with the
 Department of Physics, University of Oslo, Norway. The topic of his Ph.D. thesis is modeling and control of a
 continuum manipulator for clinical
 automation. His  main  research  interests  include  (non-)linear control, medical robotics, human-robot  interaction, and real-time systems.
 \end{IEEEbiography}

 \begin{IEEEbiography}[{\includegraphics[width=1in,height=1.25in,clip,keepaspectratio]{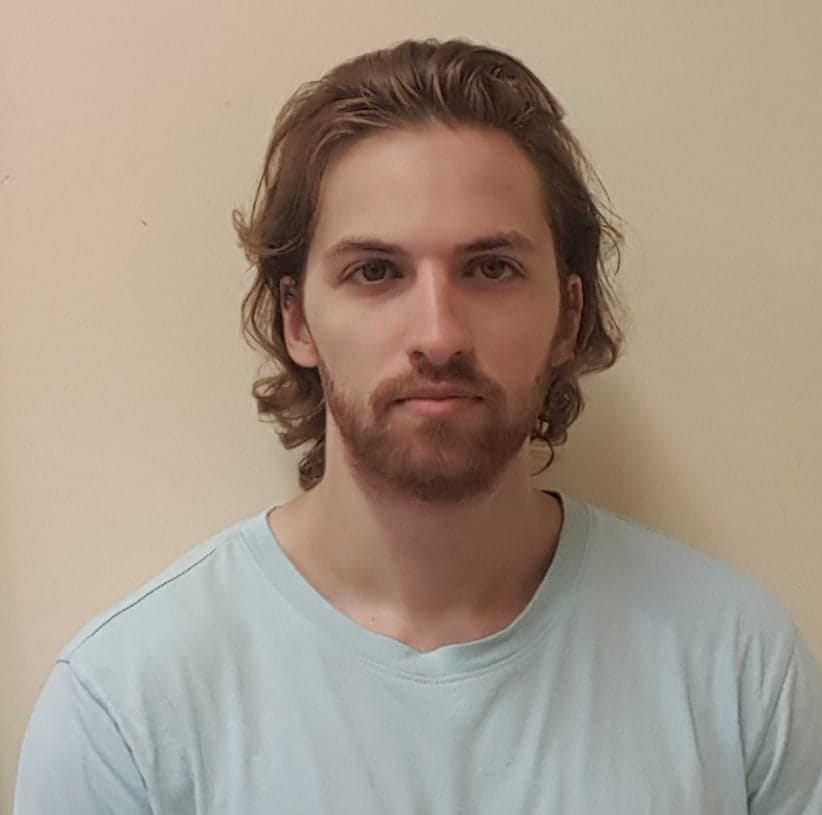}}]{Ulysse Côté-Allard}
 received the Ph.D. degree in electrical engineering from Université Laval, Québec, QC, Canada, in 2020. He is currently a researcher fellow with the Robotics and Intelligent Systems Research Group, University of Oslo, Norway. His main research interests include rehabilitation engineering, biosignal-based control, and human-robot interaction. He is a recipient of the Best Paper Award from the IEEE Systems, Man, and Cybernetics Conference and the 2022 IEEE Engineering in Medicine and Biology first Prize Paper Award.
 \end{IEEEbiography}

 \begin{IEEEbiography}[{\includegraphics[width=1in,height=1.25in,clip,keepaspectratio]{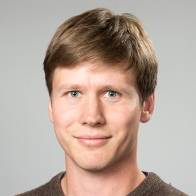}}]{Kim Mathiassen}
% received the Ph.D. degree in electrical engineering from University of Oslo, in 2017. He is currently an associate professor with the Department of Technology Systems, University of Oslo, Oslo, Norway.
 
 received a Masters degree in engineering cybernetics from the Norwegian University of Science and Technology in 2010 and a PhD in robotics from the University of Oslo in 2017. He currently works at the Norwegian Defence Research Establishment with unmanned ground vehicles and autonomous systems, and has is an associate professor at the University of Oslo teaching robotics. His current research interests span from low level robot control to high level planning and reasoning.
 \end{IEEEbiography}

 \begin{IEEEbiography}[{\includegraphics[width=1in,height=1.25in,clip,keepaspectratio]{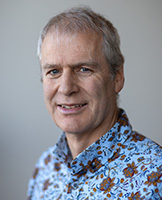}}]{Ole J. Elle}
 received the master’s degree from the Department of Production and Quality Engineering, Norwegian Institute of Technology, Trondheim, Norway, in 1990, and the Ph.D. degree in robotic surgery from the Norwegian University of Science and Technology, Trondheim, in 2004. 
 He is currently the Head of the Section for Technology Research with The Intervention Center, Oslo University Hospital, and also an Adjunct Associate Professor with the Department of Informatics, University of Oslo, Norway.
 \end{IEEEbiography}

 \begin{IEEEbiography}[{\includegraphics[width=1in,height=1.25in,clip,keepaspectratio]{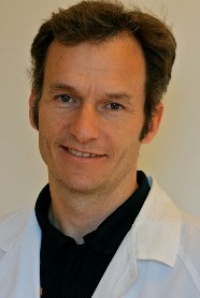}}]{ Håvard Kalvøy}
 received the Ph.D. degree from the Department of Physics, University of Oslo, Norway, in 2010. He is the head of R\&D at the Department of Clinical and Biomedical engineering at Oslo University Hospital, Norway. His research is mainly focused on medical sensor technology and the passive, electrical properties of biomaterials.
 \end{IEEEbiography}

 \begin{IEEEbiography}[{\includegraphics[width=1in,height=1.25in,clip,keepaspectratio]{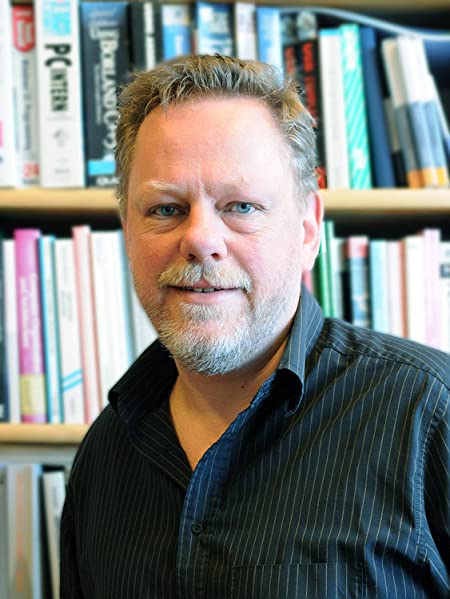}}]{Ørjan G. Martinsen}
 received the Ph.D. degree from the Department of Physics, University of Oslo, Norway in 1995 on a thesis on the electrical properties of human skin. Since then he has worked at the Department of Physics, University of Oslo, Norway from 2002 as a full professor. He also has a part time position as a senior researcher at the Department of Clinical and Biomedical Engineering at Oslo University Hospital, Norway. His research has mainly been focused on the passive, electrical properties of biomaterials. 
% He is the editor-in-chief of the Journal of Electrical Bioimpedance and a co-author of the textbook Bioimpedance and Bioelectricity Basics.
 \end{IEEEbiography}

 \begin{IEEEbiography}[{\includegraphics[width=1in,height=1.25in,clip,keepaspectratio]{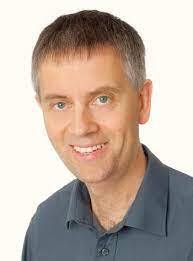}}]{Jim Tørresen}
 received the M.Sc. degree in computer architecture and design from the Norwegian University of Science and Technology, in 1991, and the Dr.Ing. (Ph.D.) degree in computer architecture and design from the University of Trondheim, in 1996.  
He is a full Professor with the Department of Informatics, University of Oslo and a Principal Investigator with the RITMO Centre for Interdisciplinary Studies in Rhythm, Time and Motion. His research interests include artificial intelligence, ethical aspects of AI and robotics, machine learning, robotics, and applying this to complex real-world applications. He is a member of the Norwegian Academy of Technological Sciences (NTVA) and the National Committee for Research Ethics in Science and Technology (NENT). 
 More details can be found at: http://jimtoer.no/.
 \end{IEEEbiography}

\end{document}